\documentclass[letterpaper,12pt]{report}
\usepackage{thesis}
\usepackage{booktabs}
\usepackage{amsmath}
\usepackage[rightcaption]{sidecap}
\usepackage{caption}
\usepackage{sidecap}
\usepackage{placeins}
\usepackage[hidelinks]{hyperref}
\usepackage{url}
\usepackage{longtable}
\usepackage{float}
\usepackage{subfigure}
\usepackage{MnSymbol}
\usepackage{graphicx}
\usepackage[ruled,vlined]{algorithm2e}
\usepackage{algorithmic}
\usepackage{xcolor}

\graphicspath{
	{MUNLogoRGB.png}
}

\begin{document}


\includegraphics[height=2.5 cm]{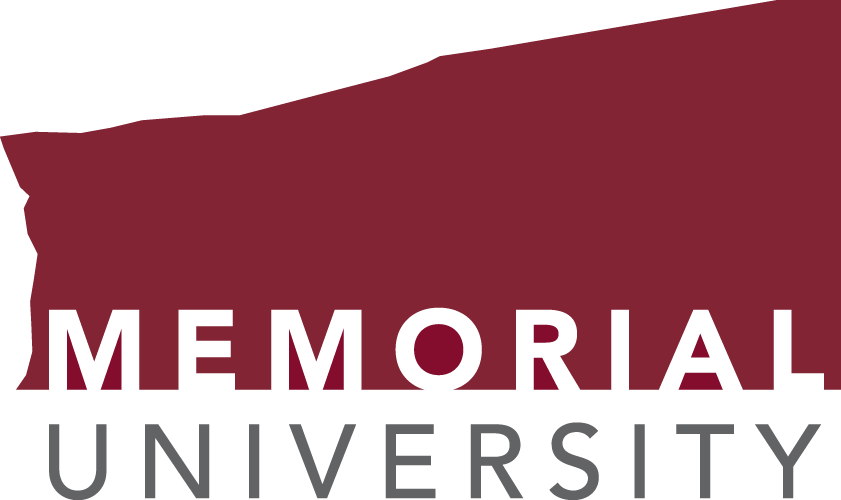}


\thesistitle
    {\Large\bfseries An Active Learning Framework with a Class Balancing Strategy for Time Series Classification}
    { Shemonto Das}
    {Master of Science}
    {Department of Computer Science}
    {February 2024}
    
\newpage
\pagenumbering{roman}

\phantomsection
\addcontentsline{toc}{chapter}{Abstract}

\textbf{\huge Abstract}
\vspace{10 pt}

Training machine learning models for classification tasks often requires labeling numerous samples, which is costly and time-consuming, especially in time series analysis. This research investigates Active Learning (AL) strategies to reduce the amount of labeled data needed for effective time series classification. Traditional AL techniques cannot control the selection of instances per class for labeling, leading to potential bias in classification performance and instance selection, particularly in imbalanced time series datasets. To address this, we propose a novel class-balancing instance selection algorithm integrated with standard AL strategies. Our approach aims to select more instances from classes with fewer labeled examples, thereby addressing imbalance in time series datasets. We demonstrate the effectiveness of our AL framework in selecting informative data samples for two distinct domains of tactile texture recognition and industrial fault detection. In robotics, our method achieves high-performance texture categorization while significantly reducing labeled training data requirements to 70\%. We also evaluate the impact of different sliding window time intervals on robotic texture classification using AL strategies. In synthetic fiber manufacturing, we adapt AL techniques to address the challenge of fault classification, aiming to minimize data annotation cost and time for industries. We also address real-life class imbalances in the multiclass industrial anomalous dataset using our class-balancing instance algorithm integrated with AL strategies. Overall, this thesis highlights the potential of our AL framework across these two distinct domains.

\phantomsection

\phantomsection
\addcontentsline{toc}{chapter}{Acknowledgments}

\textbf{\huge Acknowledgments}

\vspace{20 pt}

I would like to express my profound gratitude to my supervisors Dr. Amilcar Soares and Dr. Vinicius P. da Fonseca, for their relentless effort to the completion of my research. It was a privilege to have their expertise and support throughout the journey.

Secondly, I would thank Instrumar Limited for their support as an Industry partner for the Mitacs Accelerate Internship Program IT29753.

And finally,  I want to thank my parents, without whom I wouldn't be here, my family and friends, for their continuous support throughout the program. 

\phantomsection

\phantomsection
\addcontentsline{toc}{chapter}{Co-authorship Statement}

\textbf{\huge Co-authorship Statement}
\vspace{20 pt}

Chapter 3 is the published version of the manuscript in the journal Frontiers in Robotics and AI. 
This is the first work where we proposed our AL framework tactile robotics dataset for texture classification.
I designed and performed all the necessary experiments and tests suggested by supervisors Dr. Amilcar Soares and Dr. Vinicius P. da Fonseca. 
They both provided continuous feedback and editorial input in the initial submission phase as well as in the peer review rounds.

Chapter 4 is the accepted manuscript in the IEEE International Systems Conference.
In this phase, we worked on the industrial dataset provided by our industrial partner Instrumar Limited.
I performed all the necessary experiments suggested by supervisors Dr. Amilcar Soares and Dr. Vinicius P. da Fonseca.
Moreover, I also performed the necessary data analysis and preprocessing according to Instrumar Limited's feedback.
My supervisors, along with Instrumar, provided continuous editorial input for the initial submission.
\phantomsection

\tableofcontents
\listoffigures
\addcontentsline{toc}{chapter}{List of Figures}
\listoftables
\addcontentsline{toc}{chapter}{List of Tables}


\newpage
\pagenumbering{arabic}
\chapter{Introduction}

\section{Background}

In today's data-dependent industries, the widespread integration of sensors has led to unprecedented volumes of data requiring labeling for performing machine learning tasks.
Several areas involving human economic activities produce inherent temporal ordered information organized in time series data requiring classification.
According to Gamboa et al. \cite{gamboa2017deep}, any classification issue that uses data registered while accounting for an ordering concept can be formulated as a time series classification problem. 
Such time series data analysis has applications in many real-world industries, including healthcare, financial markets, industrial manufacturing, and environmental monitoring. 
In all these domains, using time series data is intrinsic to their operations for informed decision-making. 
For instance, in healthcare, time series data from patient monitoring devices provide valuable insights into effective disease treatment. 
Similarly, time series data on stock prices and market trends are necessary in the financial markets to enable informed investment decisions.
Time series data from sensors are used in industrial manufacturing to monitor equipment performance and production processes, ensuring operational efficiency and quality control. 
In these situations, precise predictions and classifications depend critically on efficient data labeling and management. However, labeling such data can be time-consuming and biased, especially when the data is imbalanced, where certain events or conditions are rare compared to others.


\section{Problem Statement}

Applying traditional supervised learning techniques to classify time series data encounters significant limitations regarding access to labeled data. 
The lack of annotated data for training is a significant obstacle that prevents the creation of precise classification models \cite{soares2015grasp,nasteski2017overview,zhou2018comparison,junior2017analytic,ferreira2022semi,junior2018semi}. 
In contrast to structured datasets, time series data necessitate complex and frequently expensive labeling procedures, which makes it impossible to gather a sufficiently extensive and varied annotated dataset \cite{eldele2023label}. 
The process of labeling time series data is labor-intensive and time-consuming, which further makes the process more challenging. 
It takes a lot of time and human labor to annotate large amounts of sensor readings or manufacturing process data \cite{eldele2023label}. 
This resource-intensive labeling process can cause delays, impede the development of timely decision-making systems \cite{soares2019vista,abreu2021trajectory,TOSTPub,haidri2022ptrail,haranwala2022dashboard}, and raise operating costs in real-world applications where time is frequently of the essence.

Adding to the complexity, real-life time series data frequently exhibits class imbalance, where certain events or conditions are significantly less frequent than others. 
Imbalanced datasets are prevalent in these domains due to the rarity of specific scenarios in unstructured environments for the robots to learn or defects in industrial processes. 
Traditional supervised techniques, which typically assume balanced class distributions, struggle to learn from imbalanced data effectively, often resulting in biased and inaccurate classification models. 
This thesis explores handling these limitations and enhancing the development of classification strategies using Active Learning that can address the challenges posed by the unique characteristics of time series data in industrial fiber manufacturing and tactile robotics. The fundamental idea behind Active Learning is that the learning algorithm will perform better with less training data if it has the ability to select the data from which it learns \cite{settles2009active}. 

\section{Research Goals and Questions}

This work aims to enhance the use of time series in pursuit of improved decision-making across
real-life scenarios, thus minimizing resource-intensive efforts while maximizing efficiency. 
Therefore, the goals guiding this research work are the following:

\begin{itemize}

    \item Develop an active learning pipeline for efficiently classifying time series with less annotated data.
    
    \item Design a technique to handle the imbalance within the time series data and improve the overall classification of all classes.

\end{itemize}

With these goals in mind, we therefore formulate the following Research Questions (RQ) we aim to answer through this thesis:

\begin{itemize}

    \item \textbf{RQ1.} Can active learning strategies be effectively integrated into time series classification models to minimize the need for annotated data while maintaining classification performance?
    
    \item \textbf{RQ2.} How to design an algorithm within an active learning framework to mitigate class imbalances within time series data?
    
\end{itemize}

\section{Contributions}


Our work advances the domain of time series classification through the following contributions.

\begin{itemize}

    \item The preprocessing step in Figure \ref{statistical dataframe} (Chapter 3) included in our pipeline performs necessary time interval processing based on overlapping windowing of the raw data to generate the desired features set for the machine learning models to get trained on. 
    We analyzed the effect of temporal features by evaluating the impact of the sliding window’s length and size in Table \ref{tab:f1-score-12comb} (Section 3.1 of Chapter 3). 
    This procedure is generic; it was tested in two real-world time-series datasets (e.g., robotics and manufacturing) and generates features capable of representing the underlying phenomena. 
    
    \item We implement and compare the performances of several active learning strategies with baseline strategies. 
    The results show that the active learning strategies surpass the baseline classification f-scores for all the classification models in tactile texture classification (Section 3.2, 3.3 of Chapter 3) and industrial fault classification (Section IV-A, IV-B of Chapter 4).
    
    \item We propose a class balancing instance selection algorithm that has been developed and integrated with the active learning strategies (Section 2.3 of Chapter 3 and Section III-C of Chapter 4) to analyze the effect of class imbalance on the active learning strategies and classification models. 
    We show that active learning strategies with the balancement algorithm have enhanced performance compared to those without the balancement algorithm, mainly in the early stages of the training pipeline. 
    
\end{itemize}

Chapters 3 and 4 describe all these contributions and the outcomes this thesis has produced. 
We believe our proposed framework to be generalizable in all real-life domains involving time series classification. 
Firstly, we designed the framework for classifying robotic tactile textures, and the work was published in the journal Frontiers in Robotics and AI \cite{das11active}. The accepted paper is described in Chapter 3.
Later, we used the framework for fault classification on real-world industrial data, and the work was accepted at the IEEE International Systems Conference. 
Chapter 4 is the draft of the accepted paper.



\section{Thesis Outline}

This work is presented with a thesis by articles format.  
In Chapter 2, we offer a thorough literature review of previous studies on time series classification and discuss how the novel aspects of our work progress the field, particularly in industrial fiber manufacturing and tactile robotics. 
In Chapter 3, we include the paper entitled \textit{Active Learning Strategies for Robotic Tactile Recognition Tasks} published in the journal Frontiers in Robotics and AI \cite{das11active}. 
Chapter 4 includes the draft of the paper entitled \textit{Unbalanced Fault Classification using Active Learning in Synthetic Fiber Manufacturing Process}, accepted at the IEEE International Systems Conference. 
We conclude the thesis in Chapter 5, summarizing our findings and providing the details of possible future works.
\chapter{Literature Review}

Time series classification tasks present a great need for methods to address the lack of labeled data; however, the use of AL for time series annotation still needs to be improved.  
The need for labeled time series data has emerged as a significant obstacle, especially for supervised learning techniques that rely heavily on well-annotated data for effective training and classification. 
There has been enormous research on time series classification, but integrating AL techniques to optimize and enhance classification efficacy has been rare.

Moreover, the complex nature of time series also poses many challenges to having a unique framework that generalizes any classification problem for informed decision-making. 
Due to their inherent complexities, including data imbalance, temporal dependencies, varying patterns, and irregularities, specialized methodologies are required for accurate time series classification. 
Classifying time series emphasizes the need for solid and flexible frameworks capable of handling the unique characteristics of time series data. 
Overcoming these obstacles requires innovative solutions that go beyond traditional classification techniques. 

In the following sections, we investigate the prior works in this field to determine their contribution and what can be added to our research to advance the field of time series classification. 
We start by identifying the aspects of the literature that interest us most to analyze in Section \ref{sub:analyzedAspects}. 
In Section \ref{sub:existingWorks}, we delve deeper into these works. 
Finally, we summarize the works discussed and compare this thesis with the existing literature regarding contributions in Section \ref{sub:summary}.

\section{Analyzed Aspects}
\label{sub:analyzedAspects}

Here, we highlight a few aspects that will serve as the foundation for our analysis of the contribution to the field of time series classification. 
The first aspect of our analysis involves exploring existing works that propose various methods for time series classification. 
This step aims to scrutinize whether the paper under consideration adopts a feature-based time series classification method. 
This is a crucial foundation for understanding the technical approach taken in the research, as different methods may have distinct implications for performance and efficiency in classifying time series data.
Next, we verify the focus on labeling methods; that is, does the paper address the requirement of labeling abundance of time series? 
Specifically, we seek to determine whether the paper adequately addresses the need for efficient and accurate labeling processes in the context of time series classification. 
The third aspect is whether the work uses AL techniques. 
Investigating whether the paper utilizes AL techniques provides insights into the adaptability and dynamic nature of the proposed time series classification approach.
Lastly, we identify if the paper handles class imbalance within the time series pipeline.
Identifying whether the paper tackles this issue evaluates the robustness and fairness of the proposed methodology. 
In summary, by thoroughly examining these four aspects, we aim to understand the paper's contribution to the field of time series classification comprehensively.

\section{Time Series Classification and Analysis}
\label{sub:existingWorks}

The increasing use of sensors is the driving force behind the abundance of time series in different domains, which also necessitates the need for classification to benefit decision-making \cite{schafer2017fast,spadon2022unfolding,carlini2021understanding,da2022tactile, de2022evaluating, danyamraju2023comparing}. 
Time series classification involves learning a function that maps a series into a class from a set of predefined classes \cite{nanopoulos2001feature}. 
According to Lin et al. \cite{lin2012rotation}, feature-based and whole series-based methods are two prominent time series classification techniques. 
Whole series-based methods compare the entire time series point-wise, whereas feature-based methods depend on features generated from the time series' substructure. 

In \cite{nanopoulos2001feature}, Nanopoulos et al. verified the effectiveness of feature-based classification over whole series-based classification and suggested using statistical features for time series classification. 
The major drawbacks of whole series-based classification are its sensitivity to the length of the time series and its associated noise, as it depends on the actual values of the time series for the classification task. 
The authors argued that the mentioned drawbacks can be handled by classification based on the fixed number of features extracted from the time series. This allows the classification models to learn from a fixed feature set, representing the time series' inherent pattern. 
This paper applies only to the feature-based classification aspect among all of our analyzed aspects as it focuses on classifying time series based on generated features using a multi-layer perceptron neural network. 

Susto et al. \cite{susto2018time} reviewed the existing data-driven time series classification approaches: feature-based and distance-based. 
They favor distance-based methods in the power sector that first eliminates the need to extract features from the raw time-series data to perform classification directly. 
To the authors, the preprocessing step of extracting relevant features from time series is additional complexity and can be avoided. 
They review the distance-based methods and apply them to power sector time series data. 
The distance-based methods are clustered into three groups: Pure distance-based, reduction distance-based, and parametric distance-based. The direct computation of ad hoc defined distances over raw time series is the foundation of purely distance-based techniques. 
Reduction distance-based methods calculate strategically defined distances across a condensed representation of raw time series. 
A mixture of basis signals represents raw signals in parametric distance-based techniques. 
The computation of ad hoc defined distances uses the coefficients of various representations, which are parameters. 
Though most power system applications rely on feature-based techniques, the authors review existing works using distance-based anomaly and fault detection methods. 
This paper also uses the aspect of feature-based classification.

In \cite{xing2011extracting}, Xing et al. highlight the concept of early classification for time series instead of traditional methods that extract features from the whole length of time series. 
They focus on extracting interpretable features from time series and tackle the problems associated with effective feature extraction by introducing local shapelets as features. 
They proposed a framework called EDSC that consists of two steps named Feature Extraction and Feature Selection. 
During the first stage, unique local shapelets are identified from the time series data of the training set by taking into account all subsequences up to a predetermined length and simultaneously learning robust distance thresholds. 
This procedure produces an extensive collection of unique characteristics essential for efficient time series categorization. 
In the Feature Selection step, a rule-based classifier approach was used to carefully select a small subset of local shapelets based on criteria emphasizing early classification and avoiding overfitting. 
This results in interpretability and thus increases the overall efficacy of classification. Through this two-fold framework, they align with our feature-based classification aspect.


Deep learning techniques were also investigated to enhance the effectiveness of conventional feature-based methods. Zhao et al. \cite{zhao2017convolutional} proposed a novel CNN framework for time series classification. Unlike other feature-based classification techniques, they use CNN's convolution and pooling operations to automatically find and extract the appropriate internal structure to generate deep features of the raw time series instead of using features designed by humans. 
The results suggested the performance to be better than traditional techniques. Though this work generates deep features using CNN, it differs from generating features from raw time series to fit into the models and thus does not fall under any of our analyzed aspects. 
The authors also pointed out the limitations of their work, which include the fixed length of time series during training and testing, which is mandated by the CNN architecture, and the time-consuming CNN training resulting from parameter determination through multiple experiments. 

Utilizing deeper architectures capable of comprehensively autonomously learning from annotated data makes deep learning an extremely promising methodology \cite{ismail2019deep}. 
These deep learning-based methodologies bring us to the issue of having enough labeled data and required computation resources to train such architectures. 
Gómez et al. \cite{gomez2016optical} prioritized balancing time series classification between sufficient accuracy and the best use of available resources. 
Schafer et al. \cite{schafer2017fast} mention that existing classification methods cannot scale with such a high volume of time series at an acceptable accuracy. 
Scalability and classification accuracy often remain a tradeoff for these classification methods. 
To address this, the authors have proposed a framework named WEASEL, which, using a sliding window, converts time series to feature vectors, which are then analyzed through a machine learning classifier. 
To produce discriminative features, WEASEL builds a bag-of-patterns model and uses a supervised symbolic representation to produce a discriminative feature vector. 
The process starts with taking time series data and extracting normalized windows of different lengths. 
Then, it uses the Fourier transform to approximate each window and the ANOVA F-test to determine which Fourier values best distinguish between the classes. 
These values are discretized using information gain binning to help achieve the best possible class separation. 
These features, neighboring features, and all window lengths are combined to create a unified bag of patterns. 
The Chi-squared test is used to remove features that aren't relevant. 
A quick linear time logistic regression classifier uses the highly discriminative feature vector, this process produces to classify data. 
This framework addressed the computational efficiency and accuracy and the time series feature extraction problem. 
Therefore, the paper \cite{schafer2017fast} applies to the feature-based classification aspect of ours.

In \cite{polge2020case}, Polge et al. implemented the WEASEL method in an industrial context to increase flexibility in production lines. Still, modifications have been made to the preprocessing method used in WEASEL to tackle the robustness reduction problem in dynamic settings. 
The modified preprocessing method takes the absolute values of the elements produced after standardizing time series by subtracting their mean values. 
Through the intentional shifting of series from different classes and the preservation of amplitude differences, this process aligns series within the same class, producing unique references as characteristics. 
Despite the modification, the authors are doubtful about applying such a method in real-life manufacturing processes. 
As the authors used WEASEL methods for classifying time series, this work falls under our analyzed aspect of feature-based classification. 

A Window-based Time series Feature Extraction (WTC) method has been proposed by Katircioglu-Ozturk in \cite{katircioglu2017window}. 
The efficient extraction of definitive time domain features from time series datasets was the primary goal of this work. 
The authors developed a way to generate an overall similarity score by summarizing class-dependent behaviors in successive time windows. 
To distinguish instances in a target class from those in other classes, the method finds temporal features that characterize those instances. 
The findings show that, in comparison to its shapelet-based alternatives, WTC achieves better classification performance with noticeably shorter execution times, which makes the method more robust. 
Therefore, prioritizing extracting features, this work only aligns with one of our aspects, which is feature-based classification.

Another essential challenge with time series classifications is labeling an abundance of time series instances. 
According to Woodward et al., \cite{woodward2020labelsens}, labeling is an essential step in pre-processing data that can be particularly difficult, primarily when used with real-time sensor data collection approaches that use one or more models. 
The paper \cite{woodward2020labelsens} proposes a new framework named LabelSens for real-time labeling sensor data at the point of collection, which pairs sensor data with a physical labeling method inside a tangible user interface. 
The authors advocate against offline labeling of time series as they feel it is impossible to label raw data without real-time context. 
I would argue that labeling at the point of collection may not always be feasible for many fields as the job will get tedious and expensive, potentially requiring much more human effort. 
Moreover, the work focused on labeling interfaces and experiences for annotators to label the data, but that does not address challenges related to the need for massive data labeling. The fact that human annotators need to label high amounts of data still persists despite how good the labeling interface is. 
The results show that while touch interfaces produce high labeling rates and model accuracy, users find them to be the least preferred because of the increased attention demand while using the device. 
This eventually aligns the paper with our second aspect of a labeling method.

Langer et al. \cite{langer2022visual} propose an extended version of their semi-automatic labeling tool named Gideon-TS with an active learning component integrated that splits the dataset into windows and performs unsupervised clustering to detect anomalies in them. 
Following a similarity search, these anomalous windows are identified as possible error candidates, compared to patterns from previously labeled errors, and recommended for labeling to the user. 
Once the user creates an initially labeled dataset, the labeling is continued with an active learning-based technique until the desired accuracy is reached. 
By doing this, the user can finish the process with a trained model and avoid the need to label the complete data set. 
Thus, it satisfies the requirements for including AL and the labeling method, two of our analyzed aspects.

To lower the label cost for time series classification, two important directions, namely semi-supervised learning approach and heuristics-based algorithms, are highlighted in \cite{bandyopadhyay2021automated}. 
The authors suggest automatically generating labels for unlabelled time series using a small number of representative labeled time series. 
The technique utilizes Auto Encoded Compact Sequence (AECS) for representation learning and selects an optimal distance measure. 
Through iterative self-correction and learning of latent structures, it employs a variational auto-encoder (VAE) to enhance representative time series and improve label quality, which is a critical function. 
Given the work performed, this paper applies to our analyzed aspect of a labeling method for time series. 
This methodology reduces the amount of data to be labeled but does not necessarily surpass the benchmark.

Addressing the labeling challenge associated with time series validates the need to study AL for time series classification. Peng et al.\cite{peng2017acts} have considered our aspect of using AL in the domain of time series classification. The authors introduce ACTS, a novel active learning technique based on the Nearest-Neighbor (NN) classifier for time series classification. The technique leverages shapelet discovery to find discriminative patterns in the training set and iteratively adds new examples to these patterns. A probabilistic model with instances, patterns, and labels is built to create the question selection criterion. It combines uncertainty and utility metrics to assess a time series instance's informativeness. While utility takes advantage of instance correlations, uncertainty measures classifier confidence, considering label distribution diversity and instance patterns. These metrics have been designed to solve the problem of evaluating the informativeness of time series instances and thus resulted in higher classification accuracy.  

Another nearest neighbor-based AL approach was proposed by Gweon et al. in \cite{gweon2021nearest} that works with incredibly sparse labeled data. 
This approach uses highly local information for active learning sampling, where the nearest neighbor principle is used to measure the prediction uncertainty and the utility of an unlabeled sample. 
The suggested method enables batch sampling, in which a new batch of samples is chosen from unlabeled data at each sampling iteration. Since the informativeness metric is based on distance, the result of the work demonstrates how noise variables affect the measurement of the informativeness of unlabeled instances. 
Given the limitation, the work aligns with our analyzed aspect of using AL for time series classification.

Though the above nearest neighbor-based processes can be used for classifying time series data, there are limitations. 
According to Lines et al., \cite{lines2012shapelet}, there are a lot of patterns in time series data that show the decision boundary among classes, which decreases the number of instances used in training. This makes it difficult for the nearest neighbor classifier-based method to analyze these patterns.

One of the most common problems with time series in various domains is imbalanced time series classification. 
In real-life skewed time series datasets, it becomes difficult for the classifiers to identify events represented by the minority class \cite{geng2019cost}. 
The three standard techniques for resolving imbalances are downsampling, upsampling, and class weighting \cite{buda2018systematic}. 
The original distribution of the data is changed due to these sampling techniques. When too many samples are discarded, downsampling results in subpar classifier performance, whereas upsampling causes overfitting by reusing data from the minority class. 
SMOTE and other data augmentation techniques are parametric models that are limited by computation time and do not adapt well to high-dimensional datasets \cite{dablain2022deepsmote,lusa2012evaluation}. 

Deng et al. \cite{deng2022ib} explored the aspect of imbalance in time series. The authors presented a novel technique based on Generative Adversarial Networks (GAN) called IB-GAN, which uses an imputation-balancing strategy to combine data augmentation and multivariate time series classification in a single step. 
The framework integrates seamlessly with various GAN architectures and deep learning classifiers, consisting of a triplet of generator, discriminator, and classifier models. 
The novel method of imputation and balancing uses the available training data to produce synthetic samples of higher quality for under-observed classes. 
Here, the direct feedback from classification and imputation losses improves the data quality. 
The results show that training improves classification performance in under-observed classes. 

Though CNN-based networks have been used for time series classification, most cannot classify imbalanced time series because standard networks assign the same class weights to the majority and minority classes \cite{geng2019cost}. 
To tackle this, the authors of \cite{geng2019cost} modified traditional CNN to a cost-sensitive CNN (CS-CNN), which utilizes a class-dependent cost matrix that can penalize the misclassified samples. 
Penalties are automatically modified based on the general distribution of classes and CNN's training results. 
The proposed technique changes the loss function and optimizes the process of traditional CNN, and the cost-sensitive learning assigns distinct weights to majority and minority classes. 
Even though CS-CNN produced convincing results, the work can only classify binary imbalanced time series as it lacks multi-classification considerations. This work also aligns with our aspect of addressing imbalance in time series.

Jiang et al. \cite{jiang2019gan} addressed the imbalanced time series in the industrial domain. 
They suggested a GAN-based anomaly detection method that, in particular, uses feature extraction from normal samples to train an encoder-decoder-encoder three sub-network generator. 
This makes the network independent of rare abnormal samples in an industrial context. 
Three components make up the network structure: a discriminator, a generator, and a feature extractor. 
The encoder-decoder-encoder three-subnetwork is developed during the generator design process. 
A feature extractor is intended to help shorten the training period by extracting distinctive features before supplying data to the generator. 
The authors used a discriminator to determine if input data is generated or real. As the feature extraction step is involved in the process, this work addresses two of our aspects, namely feature-based classification and data imbalancement.

\section{Summary}
\label{sub:summary}

This section summarizes the time series classification literature discussed in the previous section. 
Table \ref{review_tab} compares this thesis with the existing works discussed. 
From the table, it is evident that this thesis combines the vital aspects of time series classification, namely feature-based classification, labeling method, Al techniques, and class imbalance altogether.

\begin{table}[!htbp]
    \centering
    \caption{Comparision of this thesis with previous works}
    \label{review_tab}
    \begin{tabular}{|c|c|c|c|c|} 
        \hline
        Work  & Feature-based & Labeling   & AL    & Imbalancement \\
              & classification &  method   & inclusive     &                \\
        \hline
        A. Nanopoulos \cite{nanopoulos2001feature} & $\checkmark$ &       &       &       \\
        G.A. Susto \cite{susto2018time}            & $\checkmark$           &       &       &       \\
        Z. Xing \cite{xing2011extracting}          & $\checkmark$ &       &       &       \\
        B. Zhao \cite{zhao2017convolutional}       &            &       &       &       \\
        P. Schafer \cite{schafer2017fast}          & $\checkmark$ &       &       &       \\
        J. Polge \cite{polge2020case}              & $\checkmark$ &       &       &       \\
        D.K. Ozturk \cite{katircioglu2017window}   & $\checkmark$ &       &       &       \\
        K. Woodward \cite{woodward2020labelsens}   &            & $\checkmark$      &       &       \\
        T. Langer \cite{langer2022visual}          &        & $\checkmark$ & $\checkmark$   &   \\
        S. Bandyopadhyay \cite{bandyopadhyay2021automated}          &       & $\checkmark$ &   &   \\
        W. Jiang \cite{jiang2019gan}               & $\checkmark$   &    &    & $\checkmark$ \\
        G. Deng \cite{deng2022ib}                  &        &   &   & $\checkmark$ \\
        Y. Geng \cite{geng2019cost}                &        &   &   & $\checkmark$ \\
        F. Peng  \cite{peng2017acts}             &        &   &  $\checkmark$ &  \\
        H. Gweon \cite{gweon2021nearest}                &        &   &  $\checkmark$ &  \\
        This Thesis                              &  $\checkmark$  &  $\checkmark$  &  $\checkmark$ & $\checkmark$  \\
        \hline
    \end{tabular}
\end{table}

While extensive research has been conducted in time series classification, my investigation reveals that the utilization of active learning remains limited in this domain. 
Moreover, the existing techniques showcased certain limitations within each approach, failing to offer comprehensive solutions across all identified aspects. 
This leads to a vital necessity of having such a framework that does not limit its hypothesis to any particular characteristics of time series. 

 As we can see regarding data labeling, conventional methodologies involve manual annotation or semi-supervised techniques, which are labor-intensive, time-consuming, and may introduce biases. 
 This leads to a significant bottleneck in effectively utilizing large-scale real-life time series datasets. 
 AL techniques have shown promise in streamlining labeling efforts in various domains, yet their adoption remains limited in practical domains. 
 We believe implementing AL efficiently to annotate time series data across diverse domains has notable possibilities.

 Moreover, class imbalance in time series datasets is a problem that exists in different domains. 
 While there are several methods for addressing imbalance, such as algorithmic techniques or oversampling, these approaches are frequently inflexible and have difficulty handling imbalances unique to real-world situations. 
 More flexible and adaptable approaches are required because time series data in real-world domains often show imbalances that differ in size and distribution.

This thesis aims to not only effectively classify time series but also to develop robust and efficient classification strategies that overcome the limitations posed by particular characteristics of time series, given the ever-increasing abundance and complexity of time series data in diverse domains. 
We focus on tactile robotics and industrial manufacturing to validate our findings and results. 
The main objective is to develop a framework that can handle the complexities of real-world time series datasets while guaranteeing robustness, efficiency, and adaptability across different domains.
\chapter{Active Learning Strategies for Robotic Tactile Texture Recognition Tasks}

\section{Introduction}

Tactile perception in robots refers to their capability to detect and comprehend physical contact, pressure, and vibration information using dedicated sensors integrated into their bodies or end-effectors.
Robots can use tactile sensing to augment their perception and interaction capabilities, enabling them to perform tasks with increased precision.
Effective tactile perception allows the estimation of crucial information about the surrounding environment in various scenarios, especially under reflection, cluttered environments, challenging light conditions, and occlusion.
When relying solely on vision, robots can only identify familiar surface materials and cannot estimate their physical properties independently \cite{luo2017robotic}.
Robots can comprehend and actively engage with their surroundings when receiving static and dynamic sensory information during tactile sensing \cite{sankar2021texture}.
Tactile perception empowers robots to adapt their joints, links and reactions based on tactile feedback, resulting in improved manipulation, object recognition, and social interactions.
Tactile sensing works harmoniously with other sensing modalities like vision and proximity sensing, creating a comprehensive and versatile robot perception system. 

Touch is a vital modality while sensing the physical engagement of robots with their environments \cite{li2020review}.
Luo et al. \cite{luo2017robotic} pointed out the importance of utilizing the sense of touch to discern material properties, as it allows robots to determine essential characteristics such as surface texture (friction coefficients and roughness) and compliance, which play a vital role in object manipulation.
On the other hand, fine textures need to be slid across a surface to produce micro-vibrations that may then be analyzed and categorized.
Luo et al. \cite{luo2017robotic} emphasized the significance of incorporating the sense of touch to perceive material properties.
This capability enables robots to discern essential characteristics like surface texture, including friction coefficients and roughness, and compliance, which are crucial for effective object manipulation.
Moreover, tactile-enabled manipulation can improve robot tasks when incorporating exploratory strategies.
When analyzing and categorizing fine textures, the robot needs to slide across a surface, producing micro-vibrations that can be analyzed and categorized.

Tactile sensors such as capacitive \cite{taunyazov2019towards, pagoli2022large} and magnetic sensors \cite{yan2022tactile,bhirangi2021reskin} have been developed for robots to recognize their environment better.
In \cite{de2015touch}, the profile of a surface was recognized by dynamic touch using a robotic finger.
The motor and inertial measurement unit (IMU) feedback provided was passed through a neural network to classify shapes.
The authors of \cite{lederman1987hand} pointed out six different exploratory movements to identify the properties of an object.
Authors of \cite{de2015data} proposed a data-driven analysis for shape discrimination tasks using a robotic finger that performs the sliding movement. 
Previous works  developed an experiment and data evaluation with a sliding tactile-enabled robotic fingertip to explore textures dynamically.
The robotic fingertip's design contains a multimodal tactile sensor that makes contact with the surface\cite{lima2020dynamic, prado2021tactile, danyamraju2023comparing}.

Then, supervised machine learning models were used on this collected data to classify textures. 
Due to the nature of several textures to have essential features in different directions, this work was further developed by doing a two-dimensional exploration of surfaces \cite{lima2021classification, de2022evaluating, da2022tactile, cretu2015computational}.
Here, the authors also investigated the classification accuracy of machine learning models on tactile data from a multimodal sensing module in a dynamic exploration environment at three different velocities.
Drigalski et al. \cite{von2017textile} also classified texture on the data collected by a 3-axis force tactile sensor attached to a robot's gripper. 
Huang et al. \cite{huang2021texture} classified texture collected by a tactile sensor using a convolutional neural network.
Gao et al. \mbox{\cite{gao2020supervised}} used the BioTac sensor and auto-encoders to improve material Classification Performance. While achieving good results, the sensor used in Gao et al. \cite{gao2020supervised} work does not measure non-normal forces and changes in exploration direction, which represents a significant difference from the sensor employed in our study.
Similarly, the iCub RoboSkin in \mbox{\cite{xu2013tactile}} incorporates a non-compliant fixture, introducing substantial differences in the data characteristics compared to the information we target using the current dataset.
Numerous studies have focused on gathering more detailed tactile data for recognizing texture, resulting in abundant data for performing experiments in tactile sensing. 
Also, the data in those works are the means of teaching the robots to recognize textures.
Thus, a sufficiently labeled dataset is required for texture classification using machine learning models. 

The effectiveness of a classification model heavily depends on the data used for training.
In \cite{billard2019trends}, the authors state that robotic manipulation using machine learning in unstructured environments is computationally expensive and time-consuming. 
However, despite the advantages of having abundant data in tactile sensing, the rapidly increasing volume of data also brings specific challenges and drawbacks that require attention.
In particular, the scarcity of annotated training data has become a significant challenge for supervised learning techniques since they rely on well-annotated data for effective training. 
Labeling a large amount of data is costly and time-consuming and often necessitates the expertise of domain specialists \cite{junior2017analytic}.
To address this challenge, researchers have introduced the concept of Active Learning (AL) \cite{settles2009active,loffler2022iale}.
AL offers a solution by enabling the model to actively select and acquire the most informative data points for annotation, thus reducing the need for large amounts of labeled data.
The core concept of AL is that if the learning algorithm can choose the data from which it learns, it will perform better with less training data \cite{settles2009active}.
In scenarios where the strategy asks an expert for labels, different kinds of query strategies pave the way for deciding which instances are the most informative to be labeled. 
There have been many proposed ways of formulating such query strategies in the literature \cite{settles2009active}.
Among the query strategies, Uncertainty (UNC) sampling is the simplest and most commonly used \cite{lewis1994sequential}.
The AL system queries the expert to label the most informative instance (i.e., the instance(s) that a machine learning model is most uncertain about its class) in UNC.
There are numerous metrics to calculate uncertainty, some of which are Least Confident \cite{settles2008analysis}, Margin Sampling \cite{settles2009active}, and Entropy \cite{shannon1948mathematical}. 
Another popular query strategy is Query By Committee (QBC) \cite{settles2009active}. 
This AL strategy selects the most informative data points for labeling by querying regions in the input space where the models in a committee disagree. 
Moreover, to implement QBC, a measure of disagreement among the committee members must be established to identify the data points where the models disagree and are uncertain \cite{settles2009active}.
Finally, the Expected Model Change (EMC) \cite{freytag2014selecting} strategy, predicts the influence of an unlabeled example on future model decisions and if the unlabeled example is likely to change future decisions of the model when being labeled, it is regarded as an informative sample.

Over the last decade, studies have started incorporating AL strategies into robotics, realizing the necessity and importance of well-annotated data to teach the robots better.
In \cite{stanton2018situated}, a robot conducts unsupervised discovery to get the data it needs in dynamic settings where labeled datasets are absent.
The authors of \cite{taylor2021active} termed AL a process in which agents make decisions to collect the most relevant data to achieve the desired learning objective.
They also pointed out that informative samples are usually sparse and believed AL could fetch those samples to the robots for training, thus reducing the labeling cost and time.
Chao et al. \cite{chao2010transparent} discussed that AL is a transparent approach to machine learning as the algorithm queries an expert that provides information about areas of uncertainty in the underlying model.
In their research, they implemented AL on the Simon robot and found potential improvement in the learning process's accuracy and efficiency. 
AL has also been used for robotic grasping.
For instance, in \cite{wei2023discriminative}, authors developed a Discriminative Active Learning (DAL) framework, evaluated real-world grasping datasets, and performed better with less annotated data.
Moreover, they showed a model trained with fewer data selected by this AL framework could handle the task of real-world grasp detection.
Another framework was suggested for recognizing objects and concept acquisition in \cite{gkatzia2021s}. 
This framework's combination of few-shot learning and AL reduced the need for robot data annotation.
In their study, Sheikh et al.\cite{sheikh2020gradient} addressed AL in the context of semantic segmentation to reduce the human labeling effort of image data obtained from a mobile robot. 
Their strategies resulted in achieving higher accuracy with a reduced number of samples.

To enhance object detection with limited annotated data, the authors of \cite{xu2020active} focused on exploiting canonical views through an active sampling approach named OLIVE. 
This method selects optimal viewpoints for learning using a goodness-of-view (GOV) metric, combining model-based object detection consistency and informativeness of canonical visual features. 
Samples chosen by OLIVE, along with data augmentation, are used to train a faster regions with convolutional neural networks (R-CNN) for object detection. It is crucial to underscore that including data augmentation in their pipeline introduces an additional processing cost. 
While such a strategy effectively alleviates the burden of labeling data, it necessitates higher-cost robotic hardware for optimal performance.
Their study validates the principle of active sampling in the context of robot learning for object detection, and it aligns with our goal of minimizing annotated data required for robot training. 
In contrast to OLIVE, which uses the GOV metric, we utilize traditional AL strategies, namely UNC, QBC, and EMC, along with their uncertainty metrics to identify informative samples. 
This reduces the labeled data requirement and, like OLIVE, enhances training efficiency, improving texture recognition performance. 
Notably, the additional processing entailed by data augmentation in OLIVE is not present in our pipeline, rendering our strategy better suited for scenarios where lower-cost robotic hardware and less complex models are preferred.

Our work focuses on the potential significance of Active Learning (AL) in tactile sensing for texture recognition, particularly in the context of future robotic manipulation in unstructured environments. Recent studies \cite{Cui2021,ravichandar2020recent} highlight the significance of learning from demonstration (LfD) as the paradigm in which robots acquire new skills by imitating an expert. In addition, labeling objects from daily activities with the help of a human specialist will become fundamental for integrating robotic manipulation in unstructured environments, such as homes, universities, hotels, and hospitals. In this work, we have utilized AL to enhance the texture classification process by considerably reducing the training size of machine learning models.

Additionally, the data obtained from tactile sensors consists of time series, resulting in a large dataset to be processed. 
The volume of the collected data increases with the frequency at which it is collected, making tactile data more complex and increasing the labeling cost.
It is worth mentioning that there have been previous studies where AL has been successfully applied to time series data.
In \cite{wu2021rlad}, AL performed better using fewer samples on real and synthetic time series datasets. 
Authors in \cite{he2015active} used AL to get adequate, reliable, annotated training data for multivariate time series classification.
Often, time series data have a very imbalanced data distribution.
This may result in bias in the AL strategies while selecting the most informative data points and, thus, in the whole training process.
He et al. \cite{he2015active} have addressed the need to balance the training data among different classes while performing the classification task and believe this affects the model's generalization.
Moreover, while analyzing or classifying time series, temporal features play an essential role.
Wang et al. \cite{wang2014determination} have pointed out how rarely the time variable's effect is considered.
Their study shows that by using more temporal information, the partitioning method results in greater forecasting accuracy for sensor inputs at the feature extraction stage, and the data should be segmented into smaller segments at the feature extraction stage for sensor inputs.
For the recognition task, the authors in \cite{ma2020adaptive} mentioned that when the window size is too short, some actions may be split into numerous consecutive windows, activating the recognition task too frequently without producing high recognition results. 
According to \cite{jaen2022effects}, it is not always necessary to have large sliding windows to achieve higher performance. 
In the current paper, we aimed for a tradeoff between the information required for recognition and the cost of processing.

This study investigates the effects of using AL strategies for texture classification using tactile sensing. 
We envision a future where automated systems operate in uncertain and unstructured environments, lacking access to reliable analytic models or extensive historical datasets. 
In such scenarios, active learning and data-driven control could become paramount.
In practical scenarios, a robot capable of obtaining labels from a specialist could apply active learning strategies to select the most informative samples and obtain precise information to update its world model. 
However, without such a strategy, the sheer volume of data queries might render it unfeasible for a human specialist to assist the robotic platform effectively.

Due to the multi-class classification (12 classes) nature of our problem, we propose a class-balancing instance selection algorithm to address the imbalance issues that may arise from standard AL strategies when selecting instances for labeling. 
This algorithm is integrated into standard AL strategies (UNC, QBC and EMC) to improve the performance of texture classification models, allowing them to achieve competitive or superior performance with fewer training instances when compared to a baseline using the entire dataset.
Additionally, we use a sliding window approach to extract features from time-series data. 
We compare the effects of temporal features using two different window sizes, aiming to reduce further the training data size for robots' texture classification tasks. 
In summary, our proposed strategy combines a sliding window strategy, time-series feature extraction, and active learning with a class-balancing instance selection algorithm to reduce the number of training instances for classifying textures with supervised learning models.  
Our research significantly contributes to tactile texture recognition and robotic exploration in unstructured environments. We introduce a pipeline for data pre-processing and novel class-balancing technique within the context of active learning, addressing the challenge of imbalanced datasets in tactile texture classification. 
Our approach enhances classification performance and demonstrates the potential for robots to adapt and learn efficiently from limited human supervision, paving the way for future autonomous robotic manipulation in diverse and uncertain real-world settings. 


\section{Materials and Methods}

\begin{figure}[!ht]
   \centering 
   \includegraphics[width=.9\textwidth]{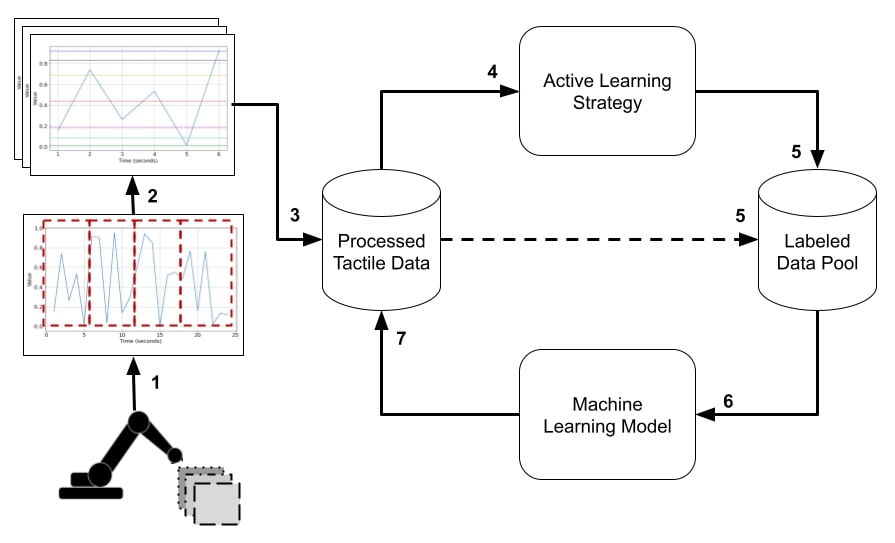}
    \caption{The pipeline for tactile data used for texture classification. 1) Data collection from exploratory movements; 2) time series data is partitioned into temporal; 3) statistical attributes extraction; 4) using AL strategies to rank instances; 5) the AL strategy selects top-ranked instances;  6) Machine-learning model built with the instances in the labeled pool; 7) classify all instances in the processed tactile data pool}
    \label{preprocessingfig}
\end{figure}

The entire pipeline developed in this work is presented in Figure \ref{preprocessingfig}. 
First, in step 1, the data used in this work was collected through exploratory movements of a tactile sensor-equipped robotics finger. 
The data is available in \cite{XYdataset, XYdatasetDIB}, and the experimental setup is explained in Section~\ref{sec:setup}.
The time series data is subsequently partitioned into more manageable temporal windows in step 2.
Extracting an array of statistical attributes from these shorter windows generates the features and processed tactile data for our machine-learning models in step 3.
After, we use AL strategies to rank instances based on how informative they are for requesting labels (step 4), and the top-ranked instances are selected by the AL strategy (step 5). 
Subsequently, these instances are attributed their appropriate labels, as already encoded in the processed data, and are consequently included in the annotated data pool (step 5).
Then, we build a machine-learning model with the instances in the labeled pool (step 6) and classify all instances in the processed tactile data pool (step 7). 
As the pipeline unfolds iteratively, steps 4 through 7 are recurrently executed until a predefined upper limit of instances is achieved, which is constrained by a pre-established budget (i.e., a maximal budget).
Ultimately, our approach's outcome is a machine learning model trained by an AL strategy and a maximal budget smaller than the total number of instances available in the original dataset.
All essential processes described in Figure \ref{preprocessingfig} of the entire pipeline are detailed in the following subsections. 

\subsection{Experimental Setup and Texture Data Collection \label{sec:setup}}

\begin{figure}[!htbp]
  \centering
  \begin{minipage}[t]{0.25\textwidth}
    \centering
    \includegraphics[height=4cm]{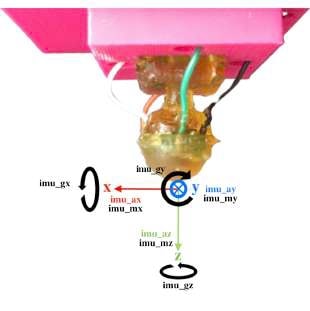}
    \caption{Multi-modal bio-inspired sensor and MARG frames of reference \cite{lima2021classification}.}
    \label{fig:sensor_tip}
  \end{minipage}%
  \hfill
  \begin{minipage}[t]{0.3\textwidth}
    \centering
    \includegraphics[height=4cm]{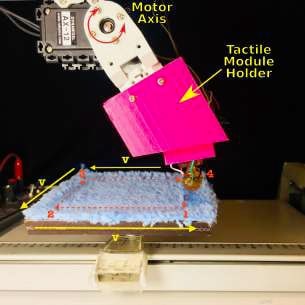}
    \caption{XY-recorder setup with a texture under exploration \cite{lima2021classification}.}
    \label{fig:sensor_mov}
  \end{minipage}%
  \hfill
  \begin{minipage}[t]{0.3\textwidth}
    \centering
    \includegraphics[height=4cm]{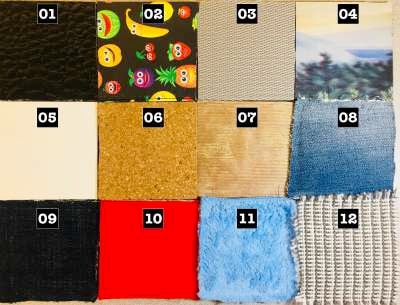} 
    \caption{The set of textures explored in the dataset \cite{lima2021classification}.}
    \label{fig:textures}
  \end{minipage}
  \label{fig:side_by_side}
\end{figure}

In this study, we used the tactile data available in \cite{XYdataset, XYdatasetDIB}, which was collected from a robotics finger from a previous study by Lima et al. \cite{lima2021classification} and represents step 1 in Figure~\ref{preprocessingfig}. 
The fingertip of the robotic finger was equipped with a fixed miniaturized tactile sensor, which was developed by Alves de Oliveira et al. \cite{de2017multimodal, de2023bioin}. 
This miniaturized sensor was also used in a previous study by Lima et al. \cite{lima2020dynamic}.
The scaled-down version of this module is depicted in Figure~\ref{fig:sensor_tip}. 

Different from \cite{lima2020dynamic}, the tactile sensing module used in this study has a rounded profile and flexible materials that enable exploratory motions, resembling the shape of a human fingertip.
During experiments, the sensor was securely held in place by an articulated robotic finger mounted on a plastic base.
The rounded form of the fingertip aims for a more straightforward exploration of 2D textures.
The module incorporates a Magnetic, Angular Rate, and Gravity (MARG) system with nine degrees of freedom, providing valuable information about the exploration vibrations. 
Additionally, the module features a base-mounted barometer that captures deep-pressure data.
Even though this unique sensor setup also allows us to record the module's deformation, in the present work, we focus on classifying textures based on the pressure data from the barometer.
Figure~\ref{fig:sensor_mov} shows the sensing module holder intended to mimic the index finger's natural probing.
The motor housing serves as a permanent representation of the intermediate phalange.
The motor axis is used to represent the intermediate-distal joint, and the tactile module holder is used to describe the distal phalange.
The distal phalange moves to bring the sensor module into touch with the surface as the intermediate-distal joint rotates.

The data \cite{XYdataset, XYdatasetDIB} used here contain readings of 12 textures
as shown in Figure \ref{fig:textures}.
More details regarding the textures are available in \cite{lima2021classification}.
The surface containing the textures was explored in a square orientation (i.e., exploring X and Y directions), as depicted in Figure~\ref{fig:sensor_mov}.
Each experimental trial involved completing a square for each texture, with 100 experimental runs conducted at three different velocities (30, 35, and 40 mm/s).
The dynamic exploration began by lowering the fingertip until it made contact with the textured surface.
A small torque was applied to the fingertip to ensure consistent contact throughout the experiment.
An empirically chosen pressure threshold determined when the fingertip torque would end.
The total run time for each experiment was 12 seconds, with 3 seconds required to travel each side of the square.
The dataset uses a unique compliant multi-modal tactile sensor comprising a deep pressure sensor and inertial measurements, which provides the interesting ability to detect micro-vibrations and changes in pressure non-normal to the surface while changing directions, as performed in this dataset, which is expected from robots exploring daily textures.
More details regarding the experimental setup and data collection approach can be found in \cite{lima2021classification}.

\subsection{Data Preprocessing and Feature Extraction}

Based on the findings of \cite{lima2021classification}, the barometer feature achieved the best classification results, exhibiting 100\% accuracy for most texture classification cases. 
Therefore, we decided to rely solely on the barometer sensor to conduct our experiments to investigate dimensionality reduction.
For our work, we utilized 1200 samples (corresponding to 12 textures) collected solely using the barometer at a velocity of 30 mm/s. 
Each sample contains data collected at 350 Hz while exploring the respective textures.
Subsequently, we applied our preprocessing pipeline to these raw samples. 
The pipeline involves using an overlapping sliding window with a configurable window size, where the overlap was set to 50\% of the window size. 
We believe setting higher overlapping thresholds could result in excessively similar examples within the sequence, potentially reducing diversity in the training set and increasing the processing time required to transform readings into machine-learning-ready examples. 
The consequence of such higher overlapping thresholds is that it could pose challenges in resource-constrained environments.
We experimented with two window sizes (3 and 6 seconds) to explore the impact of temporal features on our texture classification task.
The values of 3 and 6 seconds used in this work were set mainly because of the setup of how the data was collected. A sliding window of 3 seconds corresponds to 25\% exploration, while 6 seconds corresponds to 50\% exploration. 
In the experiments, we aim to capture distinct dimensions of exploration within these time slices, and extending the exploration beyond 6 seconds would lead the module to revisit dimensions already explored in the initial 3 seconds, resulting in redundant data collection
For each window, the pipeline generates a scaled instance (i.e., a MinMax scaler with values ranging from 0 to 1) comprising 11 statistical features, including mean, median, variance, skewness, standard deviation, quantiles (10, 25, 76, 90), min, and max. 
In this way, a resultant statistical features data frame is generated for each time window on which the machine learning models are trained. 

\begin{figure}[!ht]
   \centering 
   \includegraphics[width=1.\textwidth]{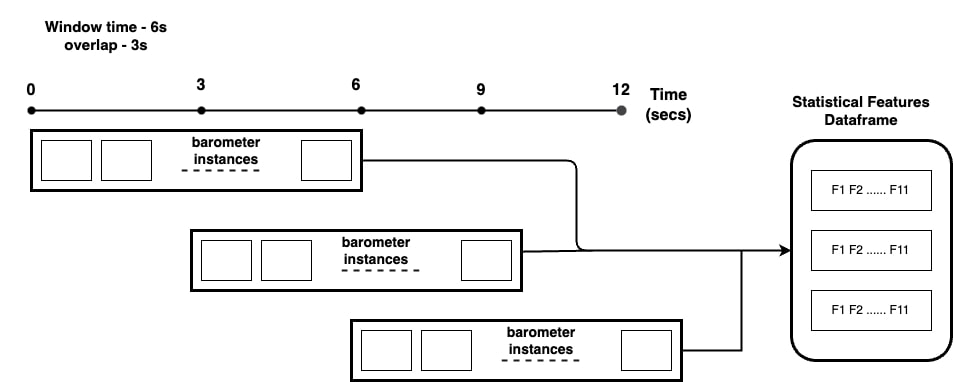}
    \caption{Window-based statistical features extraction for the 6 seconds with 3 seconds overlap.}
    \label{statistical dataframe}
\end{figure}

This generation of features uses the window strategy depicted in Figure \ref{statistical dataframe} and described below.
Given as input a time window of 6 seconds and an overlap of 3 seconds, the first instance $i_1$ generated from one experiment would start on time $t = 0$ and end time $t = 6$. 
From all the barometer data collected from this time window, we extract the 11 statistical features and add such features to a data frame, storing the texture label of that particular instance. 
For the second instance of the same experiment with a texture, we move the window by 3 seconds (i.e., start on time $t = 3$ and end time $t = 9$) and repeat the process of extracting features and adding a label to the texture. 
This entire procedure is repeated until the end of each experiment with the textures. 
Such an approach generated two datasets, one for a 3 seconds window with 50\% overlap consisting of 6718 instances, and another for a 6 seconds window with 50\% overlap and 2878 instances. 
The processed data (step 3) referred to in this subsection is the processed tactile data discussed in Figure \ref{preprocessingfig}. 





\subsection{Active Learning Strategies and Class-Balancement Instance Selection Algorithm}

AL strategies aim to select the most uncertain instances classified by a model to be labeled by an expert so that the model can learn better with less data.
Our research compares three standard AL approaches, Uncertainty sampling (UNC) \cite{settles2009active},  Query by Committee (QBC) with bagging \cite{settles2009active} and Expected Model Change (EMC). 
The UNC strategy selects for labeling the most informative instances from an unlabeled data pool in an iterative manner. 
It measures the model's uncertainty about its predictions using metrics such as entropy, margin sampling, or least confident predictions \cite{settles2009active}. 
The labeled data is then used to improve the model through training updates. 
QBC with Bagging is an active learning strategy combining QBC with a bagging ensemble. 
It generates diverse models using bootstrapped sampling and identifies uncertain instances by assessing disagreement among the committee of models. 
These instances with a high degree of uncertainty are labeled, and the ensemble is continually updated with the new labeled data. 
The EMC strategy identifies instances expected to induce the maximum positive change in the model's performance when labeled. This approach estimates the potential improvement by evaluating the impact of each unlabeled instance on the model's behavior, and the instance with the highest expected positive change is selected for labeling.

In our experiments, we split 80\% of our processed data as a training set and 20\% into a test set for validation purposes.
In our AL strategy, the training set is used as an unlabelled pool (i.e., the Processed Tactile Data in Figure \ref{preprocessingfig}) from which the most uncertain labeled instances are queried for being labeled.
The labels are extracted from the raw data obtained by the robotic arm with tactile sensors, and therefore, we do not involve a user annotating the data in our experiments.
Once an instance is labeled, it is added to the pool of labeled data on which a machine-learning model can be trained.
As AL is an iterative process, we fix a maximum annotation budget (i.e., a maximum number of instances that can be queried throughout the process) and a step size that defines the number of instances to be labeled at each batched query we make on the unlabeled pool.
The whole experiment was controlled and repeated 20 times. 
We used 20 seed values for experimental reproducibility. 
The seed generation procedure used the decimals of $\pi$ selected four by four (i.e., $seed_1 = 1415$, $seed_2 = 9265$, etc).
Such a decision avoids an arbitrary choice of seeds. 
At the first iteration of the AL strategies, instances are queried randomly (step 4 in Figure \ref{preprocessingfig}). 
After we obtain the first subset of labeled training data, and then the AL strategy is executed (step 5 in Figure \ref{preprocessingfig}), ranking and selecting the most informative instances to be labeled.
For UNC, we used Least Confidence \cite{settles2008analysis} as a metric for evaluating the uncertainty of instances defined in Equation \ref{eq:least_confidence}.
Using Equation \ref{eq:least_confidence} will query the instances with the lowest confidence and the highest uncertainty for getting labeled.
Here, $y^*$ is the most likely class label assigned by a machine learning model, and $\theta$ is the parameters of the model.

\begin{equation} \label{eq:least_confidence}
    \phi^{\text{LC}}(x) = 1 - P(y^* | x; \theta)
\end{equation}

While implementing QBC with bagging, we have created a committee with base models trained on different subsets of labeled data using bagging. 
The committee of models collectively predicts the labels for unlabeled instances or candidate samples. 
Each model in the committee provides its prediction for each instance, and the level of disagreement among the committee members is measured to assess the uncertainty or informativeness of each candidate sample.
For our case, we have used vote entropy \cite{settles2008analysis}, shown in Equation \ref{eq:vote_entropy}, to calculate the disagreement among the committee members.
Using Equation \ref{eq:vote_entropy}, the most informative samples were queried from the unlabelled pool and added to the labeled pool for training.
In Equation \ref{eq:vote_entropy}, $V(y_{t,m})$ is the number of votes a particular label n receives from the committee of classifiers, C and T is the total number of instances.

\begin{equation}\label{eq:vote_entropy}
    \phi^{VE}(x) = -\frac{1}{T} \sum_{t=1}^{T} \sum_{n=1}^{N} \frac{V(y_{t,n})}{C} \frac{\log V(y_{t,n})}{C} 
\end{equation}

We instantiated the Expected Model Change (EMC) through Equation \ref{eq:emc}, which quantifies the performance differential resulting from the incorporation of an unlabeled instance. This instance is assigned a potential label that maximizes the expected impact on performance. 
In Equation \ref{eq:emc}, $x$ is the instance to be evaluated, $y_i$ are all labels possible in a given data set, and $M$ is a performance metric. 
In this work, we used the f1-score as our performance metric to calculate the EMC.

\begin{equation} \label{eq:emc}
 \phi^{MC}(x) = \max_{\{y_i \in [l_1, l_k]
\}}(M(x, y_i) - M) 
\end{equation}


This paper presents a novel class-balancing instance selection algorithm that effectively addresses the class imbalance issue in the data. 
The main objective is to query the unlabeled instance pool in a manner that ensures a balanced representation of classes in the labeled data. 
This approach aims to prevent bias in the model's classification performance, which may occur when certain classes have a disproportionate number of instances selected by an active learning (AL) strategy.
The core idea behind our algorithm is to modify the class frequencies in the training set by inverting their occurrences. Consequently, we prioritize selecting more instances from classes with fewer instances in the training set while reducing the number of instances chosen from classes with more instances. To illustrate, consider a problem with four classes, where the instance frequencies in the training set from an AL strategy are $f\_class_1 = 0.4$, $f\_class_2 = 0.15$, $f\_class_3 = 0.2$, and $f\_class_4 = 0.25$. In the next round of instance selection to be labeled by an expert, we would invert these frequencies, leading to the selection of 15\% of instances from class 1, 40\% from class 2, 25\% from class 3, and 20\% from class 4.
By employing this strategy, we endeavor to achieve a balanced training set. However, it is essential to acknowledge that perfect balance cannot be guaranteed, given uncertainties associated with the class label of the selected instance, which the expert will ultimately resolve.
In summary, our proposed class-balancing instance selection algorithm offers a promising approach to address a class imbalance in active learning, mitigating potential bias and enhancing the overall classification performance by striving for a more equitable representation of classes in the training data.

A pseudo-code of our strategy is given in Algorithm \ref{alg:compute-inverted-frequency}. 
In summary, the method selectively adjusts the class frequencies to emphasize classes with lower representation while reducing the prevalence of classes with higher representation. 
This process enhances the balance in class distribution, which is crucial for the better performance of various machine-learning algorithms. 
In line 1, the algorithm starts by computing the frequency of each class in the training data. The $bincount()$ function is used to count the occurrences of unique labels in $current\_train\_y$, resulting in an array named $freq$.
In line 2, the frequency values obtained in the first step are normalized to a range between 0 and 1. 
Each frequency value is divided by the number of examples in the training set (len($current\_train\_y$)) to yield a new array named $norm\_freq$. 
To achieve an inverted probability distribution, the $norm\_freq$ array is sorted in descending order using the $sort()$ function, and the $[::-1]$ slicing is applied to reverse the order of the sorted array. The resulting array is named $sorted\_freq$ (line 3).
Next, to determine the mapping that would invert the probabilities, an auxiliary array named $sorted\_indices$ is created (line 4). This array stores the indices that would sort the $norm\_freq$ array in ascending order when sorted twice using the $argsort()$ function. 
Then, a loop (lines 5 to 7) is executed, iterating over each element in $sorted\_indices$. 
For each index $i$, the corresponding value in $sorted\_freq$ is assigned to the $inverted\_freq$ list at the position $i$.
After all iterations, the algorithm has constructed the $inverted\_freq$ list, which now contains the inverted frequency values for each class in the training set that is returned as the result of the algorithm in line 8. This algorithm was integrated with both AL strategies, and its impact on AL strategies is presented in the Results section. 

     

\begin{algorithm}
\caption{Compute Inverted Example Frequency}
\label{alg:compute-inverted-frequency}
\hspace*{\algorithmicindent} 
\textbf{Input:} $current\_train\_y$: all label values used in the current training set\\
\hspace*{\algorithmicindent} \textbf{Output:} $inverted\_freq$: The inverted frequency of the class distribution 
\begin{algorithmic}[1]

\STATE $freq \gets \text{bincount}(current\_train\_y)$
\STATE $norm\_freq \gets \frac{freq}{\text{len}(current\_train\_y)}$
\STATE $sorted\_freq \gets \text{sort}(norm\_freq)[::-1]$
\STATE $sorted\_indices \gets \text{argsort}(\text{argsort}(norm\_freq))$
\FOR{$i$ \textbf{in} $sorted\_indices$}
    \STATE $inverted\_freq[i] \gets sorted\_freq[i]$
\ENDFOR
\RETURN $inverted\_freq$
\end{algorithmic}
\end{algorithm}


\subsection{Classification Models and Evaluation Metric} 

In our active learning approach, we can integrate any classifier as a model to predict the instance classes (step 6 in Figure \ref{preprocessingfig}). 
For our experiments, we opted for widely-used classifiers available in the $sklearn$ package, including Decision Trees \cite{rokach2005decision}, Extra Trees \cite{geurts2006extremely}, XGBoost \cite{chen2016xgboost}, and Random Forest \cite{breiman2001random}.
A decision tree recursively splits the dataset into subsets based on the most significant feature at each level of the tree, making decisions along the way until it reaches a decision at the leaf nodes. 
A Random Forest is an ensemble model that collects weak learners' predictions and aggregates their individual predictions to produce a final and more robust forecast. 
The Extra Trees (i.e., Extremely Randomized Trees) model is also an ensemble algorithm that builds multiple decision trees, but it introduces additional randomness on the section of the feature and thresholds in the tree-building process. 
Finally, the XGBoost (i.e., Extreme Gradient Boosting) is an ensemble model that is trained sequentially and uses the gradient descent optimization technique to minimize a loss function while adding weak learners to the ensemble.
Throughout our experiments, we used the base version of these classifiers from the scikit-learn package \cite{pedregosa2011scikit}, employing their standard hyperparameters detailed in Table \ref{tbl:hyper}.
All ensemble techniques (i.e., Random Forest, Extra Trees, XGBoost) used a Decision Tree as the weak learner.

\begin{table}[!ht]
\centering
\begin{tabular}{@{}ll@{}}
\toprule
\textbf{Model} & \textbf{Hyperparameters}                                                                                                       \\ \midrule
Decision Tree  & \begin{tabular}[c]{@{}l@{}} \textit{split criterion:} gini index \\ \textit{minimal samples to split:} 2\end{tabular}                                \\ \midrule
Extra Trees    & \begin{tabular}[c]{@{}l@{}}\textit{split criterion:} gini index\\ \textit{minimal samples to split:} 2 \\ \textit{number of estimators:} 100\end{tabular}  \\ \midrule
Random  Forest & \begin{tabular}[c]{@{}l@{}}\textit{split criterion:} gini index \\ \textit{minimal samples to split:} 2  \\ \textit{number of estimators:} 100\end{tabular} \\ \midrule
XGBoost        & \begin{tabular}[c]{@{}l@{}}\textit{learning rate:} 0.3\\ \textit{max depth:} 6 \\ \textit{number of estimators:} 100 \end{tabular}                                                   \\ \bottomrule
\end{tabular}
\caption{Models' hyperparameter settings.}
\label{tbl:hyper}
\end{table}

This choice ensures the classifiers are utilized in their default configurations, allowing for fair and consistent comparisons during our evaluations.
By employing these standard classifiers, we aim to comprehensively assess their performance in the context of active learning for texture classification tasks.
In this work, we chose to employ shallow models because one of the primary considerations is that processing efficiency is paramount in the context of embedded systems.
Therefore, using deep models can introduce significant computational burdens that may not align with the expected processing capabilities of robotic platforms operating in unstructured environments.

Our work adopted a different evaluation metric than the approach used in \cite{lima2021classification}. 
Instead of using accuracy, we utilized the f1-score as our evaluation metric.
The choice of f1-score as the evaluation metric is supported by \cite{sokolova2009systematic}, where it was highlighted that accuracy, although commonly used for classification performance evaluation, may not adequately address class imbalance issues. 
While our data may be balanced, our active learning (AL) strategies could still suffer from the effects of class imbalance while selecting instances from different classes in its iterative process. 
Hence, we deemed it necessary to consider the f1-score as our evaluation metric.
Unlike accuracy, the f1-score considers precision and recall, making it a more balanced and informative evaluation metric, especially in scenarios involving imbalanced class distributions. 
By incorporating both precision and recall, the f1-score provides a comprehensive assessment of the classifier's performance, ensuring a more robust evaluation of our texture classification task.
The models produced at this step are iteratively trained in the labeled data pool. 
At every training cycle (steps 4 to 7 in Figure \ref{preprocessingfig}), the number of labeled data inside the pool is increased and we record the model's achieved performances for an in-depth evaluation of our entire proposed pipeline.

\section{Results}

In this section, we conduct a comprehensive analysis of the performance of our AL strategies using four different classifiers, employing f1-score as the evaluation metric.
To establish the baseline performance of each classifier considered in this work, we train and test them using repeated (4 times) 5-fold cross-validation. 
The baseline f1-score values are a reference for comparing all combinations of machine learning models and AL strategies used in this work since they represent an average performance of the models using the entire dataset for building a model. 
It is essential to point out that the baseline value is not a ceiling value for the problem discussed in this paper since changing the training data will likely change the decision surface of a machine learning model and, as a consequence, there is a possibility of obtaining a better model than simply using more data to train a model.
In this section, we start by exploring the influence of temporal features in the texture classification task by comparing the results obtained using different window sizes (e.g., 3 and 6 seconds) and window overlaps (e.g., 75\%, 50\%, 25\%).
Furthermore, we investigate the impact of our class-balancing instance selection algorithm on the AL strategies by generating results both with and without such an algorithm. 
This comparison allows us to assess the effect of the class-balancing approach on the performance of the AL strategies.
The figures in this section present the number of instances queried for labeling at each Step (x-axis) against the evaluation metric f1-score values (y-axis).
For each step in the querying process, we present two box plots. 
The blue box plot shows the performance of the AL strategy (e.g., UNC or QBC) with the class-balancing algorithm, while the red box plot presents the performance without the algorithm.
We apply a sliding overlapping window with several durations and overlaps on the raw data to prepare the data for analysis. 
This preprocessing pipeline generates processed datasets of statistical features, as discussed in Section 2.2.
The AL strategies are then executed on these processed datasets, allowing us to examine their effectiveness in texture classification.

\subsection{Sliding Window Size and Overlap Percentage Analysis}

We start by assessing the influence of the size and duration of the sliding window.
We have tested several combinations of parameter values to understand how these values affect our experiments.
The sliding window size was tested for 1, 3, 6, and 9 seconds, as these values are sufficient to test robots exploring the axis of objects in one (e.g., 1 and 3 seconds) or two directions (6 and 9 seconds). 
We also tested overlaps of 75\%, 50\%, and 25\%, aiming at generating more or less data for the machine learning models and evaluating how their performances would be affected by those different amounts of training data.
We tested all combinations of these parameter values (4 sliding window sizes and 3 overlapping values).
A sliding window of 9 seconds with an overlap of 25\% cannot be obtained in our experiments since the robot exploration is of at most 12 seconds, and a 25\% overlap would start the next window with 6.75 seconds and therefore could not obtain a total window size of 9 seconds (i.e., starting at 6.75 seconds and ending at 15.75 seconds).
Therefore, we test 11 combinations of sliding window sizes and overlaps. 
The details of the datasets generated by all these 11 combinations can be seen in Table \ref{tab:instances-gen}.
Finally, we have used a repeated (4 times) 5-fold cross-validation for all the machine learning models (e.g., Decision Tree, Extra Trees, Random Forest, and XGBoost), and the average f1-scores are reported in Table \ref{tab:f1-score-12comb}.

\begin{table}[!ht]
\centering
\begin{tabular}{@{}lccc@{}}
\toprule
\multicolumn{1}{c}{\multirow{2}{*}{\textbf{Sliding window size}}} & \multicolumn{3}{c}{\textbf{Overlapping percentage}} \\  
\multicolumn{1}{c}{} & \textbf{75\%} & \textbf{50\%} & \textbf{25\%} \\ \midrule 
1 second & 53,997 & 27,598 & 17,999 \\
3 seconds & 15,599 & 8,399 & 5,999 \\
6 seconds & 5,999 & 3,599 & 2,400 \\
9 seconds & 2,400 & 1,200 & - \\ \bottomrule
\end{tabular}
\captionsetup{justification=centering}
\caption{The total number of instances generated by our sliding window procedure when using the 11 sliding window size and overlap combinations.}
\label{tab:instances-gen}
\end{table}

\begin{table}[ht!]
\centering
\small
\begin{tabular}{@{}lllllllllllll@{}}
\toprule
\textbf{Window size} & \multicolumn{3}{c}{\textbf{1 second}} & \multicolumn{3}{c}{\textbf{3 seconds}} & \multicolumn{3}{c}{\textbf{6 seconds}} & \multicolumn{3}{c}{\textbf{9 seconds}} \\ 
\textbf{Overlap} & \multicolumn{1}{c}{\textbf{75\%}} & \multicolumn{1}{c}{\textbf{50\%}} & \multicolumn{1}{c}{\textbf{25\%}} & \multicolumn{1}{c}{\textbf{75\%}} & \multicolumn{1}{c}{\textbf{50\%}} & \multicolumn{1}{c}{\textbf{25\%}} & \multicolumn{1}{c}{\textbf{75\%}} & \multicolumn{1}{c}{\textbf{50\%}} & \multicolumn{1}{c}{\textbf{25\%}} & \multicolumn{1}{c}{\textbf{75\%}} & \multicolumn{1}{c}{\textbf{50\%}}  \\ \midrule
Decision Tree & 51.21 &  51.62 & 55.64 & 70.78 & 72.82 & 75.33 & 80.90 & 80.90 & 79.43 & 87.82 & 88.55 \\
Extra Trees &  64.73 & 63.66 & 67.04 & 82.99 & 83.89 & 84.63 & 90.64 & 89.75 & 88.36 & 94.40 & 93.80 \\
Random Forest &  64.24 &  63.66 & 66.80 & 82.31 & 83.18 & 83.92 & 89.51 & 88.83 & 87.54 & 92.81 & 91.03  \\
XGBoost &  65.50 & 64.87 & 67.41 & 83.50 & 84.01 & 84.29 & 90.56 & 89.45 & 87.56 & 93.60 & 92.63  \\ \bottomrule
\end{tabular}
\captionsetup{justification=centering}
\caption{Average f1-score values for all 11 combinations of sliding window sizes and overlaps.}
\label{tab:f1-score-12comb}
\end{table}

The results presented in Table \ref{tab:f1-score-12comb} show that a window of 1 second is not enough to distinguish the textures. 
This means that using a window of 1 second cannot extract meaningful features to classify the textures since the f1-scores are in a range between 51\% to 67\%.
Higher performances are achieved when the robot starts exploring the texture for 3 to 9 seconds.It is essential to point out that the objective of our work is to learn with as little data as possible, and we relied on Table \ref{tab:f1-score-12comb} to decide on the proper values for the window size and overlap. We discarded 9-second windows mainly because the number of instances to be chosen for an AL strategy would be low (i.e., 1,200 to 2,400 instances) when compared with other setups, and this could decrease the advantage of selecting high-yield instances from the unlabeled pool. In addition, a robot should explore the surface as little as possible to predict the proper label for the texture and using 9 seconds would make the robot explore a direction (i.e., x or y axis) a second time. It is important to highlight that higher performance values with longer windows are expected since the robot would explore them for a longer time on the surface and have a higher chance of correctly classifying the texture. Finally, we discarded the overlapping windows of 25\% and 75\% since the differences between them, when compared to a 50\% window, were not statistically significant for 3 and 6 seconds windows. Therefore, we decided to provide a balance between performance and a reasonable pool of instances to be given to the AL strategies, which is obtained by using the 50\% overlap. As a result from this experiment, we conclude that using windows of 3 and 6 seconds with an overlap of 50\% is a reasonable choice (i.e., obtaining higher performances while maintaining a reasonable pool of instances for the AL strategies) of parameters for the data set used in our experiments and, therefore, we use this result to proceed with our AL strategies analysis.

\subsection{Uncertainty Sampling Strategy on 3 and 6 seconds windows}

\begin{figure}[!ht]
   \includegraphics[width=.9\textwidth]{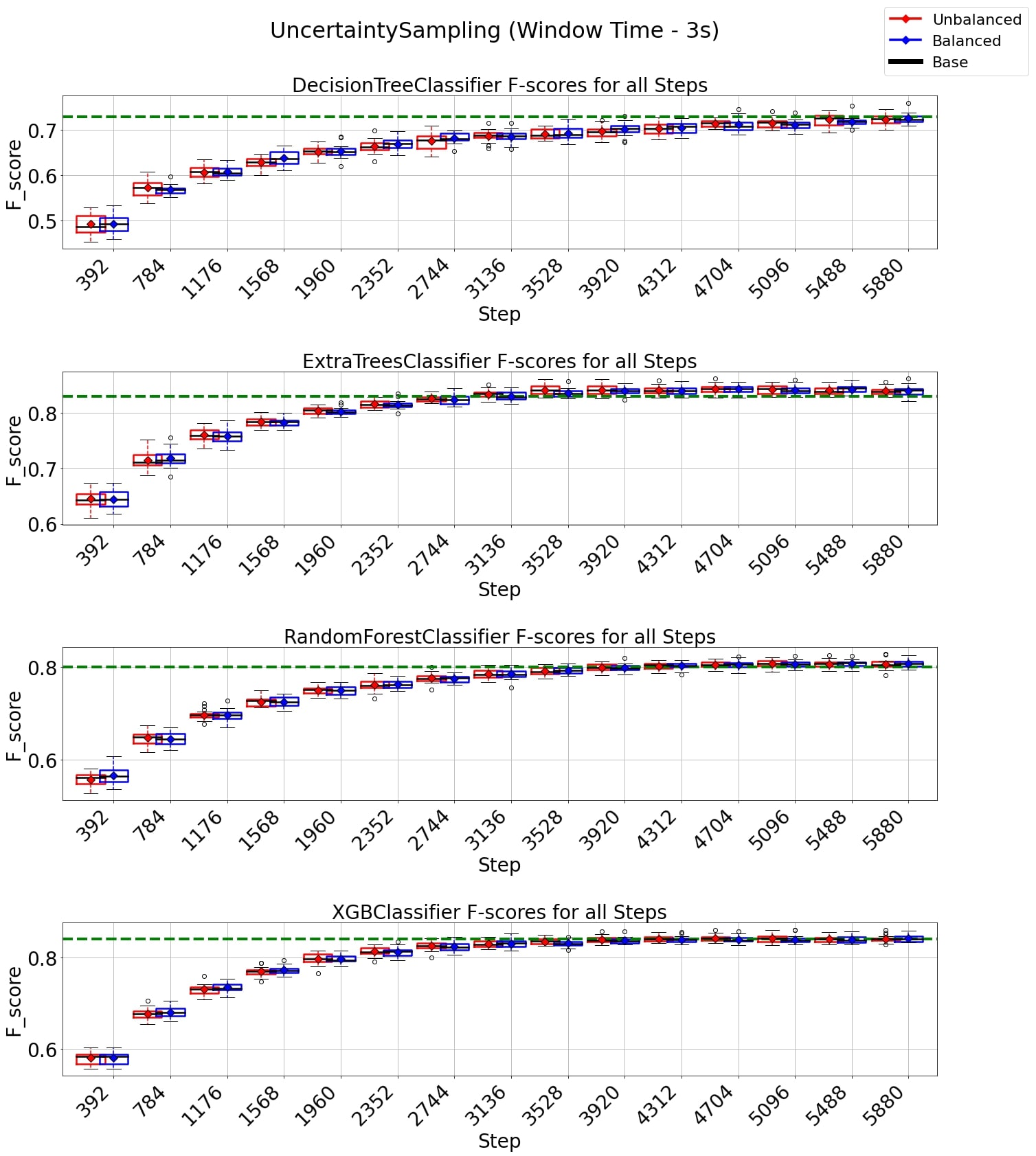}
    \caption{f1-score plots of the uncertainty sampling strategy for the 3-second window.}
    \label{unc3s}
\end{figure}

In this experiment, we adopted the uncertainty sampling (UNC) strategy to minimize the number of instances requiring expert annotation and to enhance texture classification performance.
Figure \ref{unc3s} illustrates the results obtained using a window size of 3 seconds and an overlap of 50\%. 
All classifiers using the UNC strategy managed to achieve and surpass their respective baseline f1-score values within an annotation budget of approximately 5880 out of a total of 8399 instances (equivalent to 70\% of the data).
This finding suggests that we could reduce the training set for these classifiers by 30\% while achieving comparable or even improved performance compared to utilizing the entire dataset for training. 
Notably, for the ExtraTrees, Random Forest, and XGBoost classifiers, the total budget could be reduced to only 3920 instances (equivalent to 47\% of the data) to achieve baseline performance.
It is also noticeable that UNC with the class-balancement algorithm performed similarly to the standard UNC algorithm, and therefore, no substantial gains were observed by using our class-balance instance selection algorithm.

\begin{figure}[!ht]
   \centering 
   \includegraphics[width=.9\textwidth]{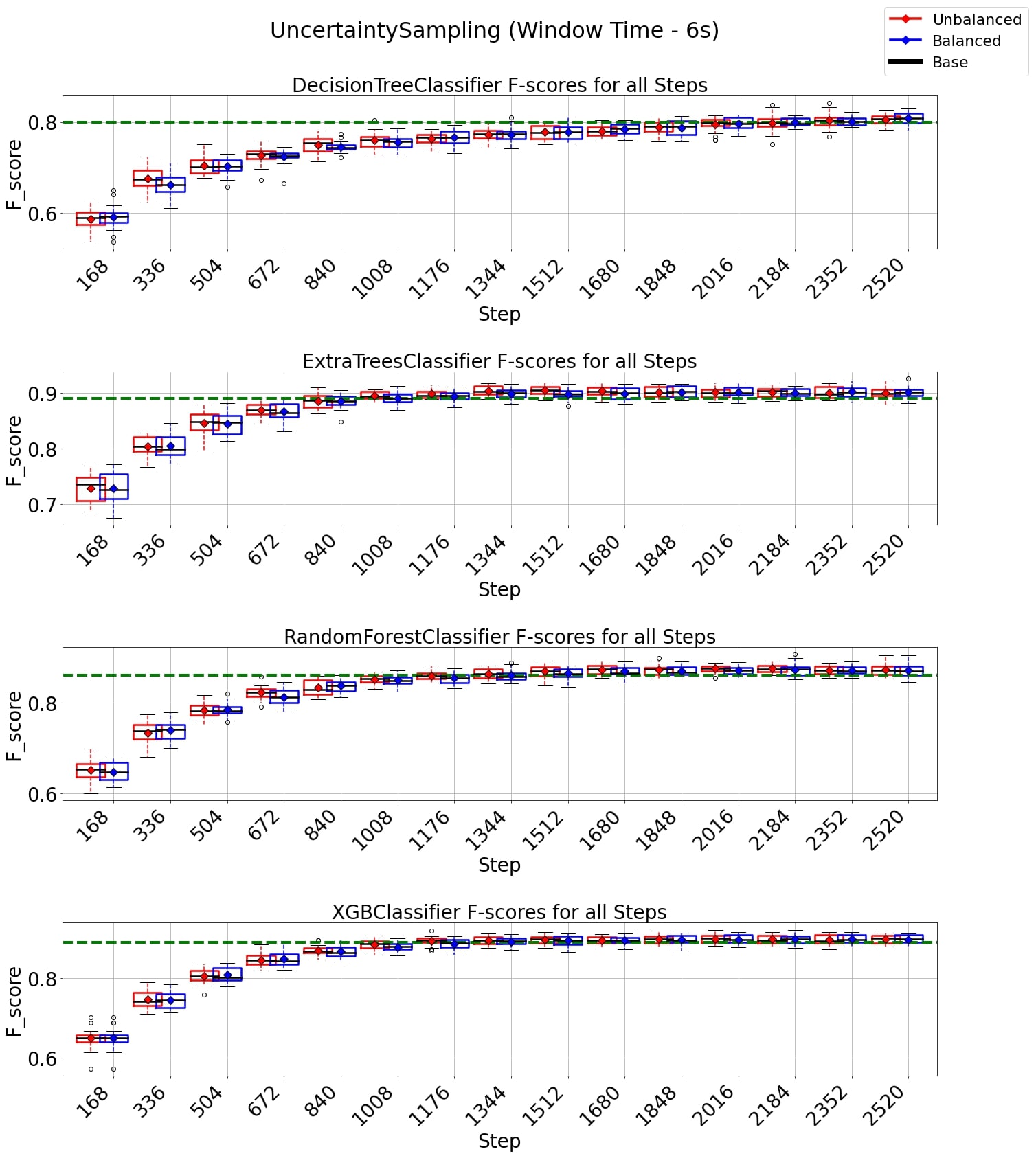}
    \caption{f1-score plots of the uncertainty sampling strategy for the 6-seconds window.}
    \label{unc6s}
\end{figure}

In Figure \ref{unc6s}, we present the results obtained using the uncertainty sampling (UNC) strategy with a window size of 6 seconds. 
For this configuration, all classification models achieved performance exceeding the baseline with an annotation budget of 2016 instances (equivalent to 56\% of the total 3599 instances in the dataset).
This finding indicates that a substantial reduction of 44\% in instances is possible when employing the UNC strategy to train machine learning models in this texture classification task without compromising performance. 
Again, in the case of Extra Trees, Random Forest, and XGBoost a further reduction for 1176 instances (33\% of all training data) is enough to surpass the baseline, indicating a reduction of 67\% of the training data is possible and still achieve the baseline value. 
Furthermore, when combined with the class-balancing instance selection algorithm (shown in blue), we observed that the UNC strategy demonstrated performance similar to the standard UNC algorithm (shown in red). 
This suggests that no significant improvements were achieved by balancing the selection of the next round of instances when using UNC as the active learning strategy. 


\subsection{QBC Strategy on 3 and 6 seconds windows}

\begin{figure}[!ht]
   \centering 
   \includegraphics[width=.95\textwidth]{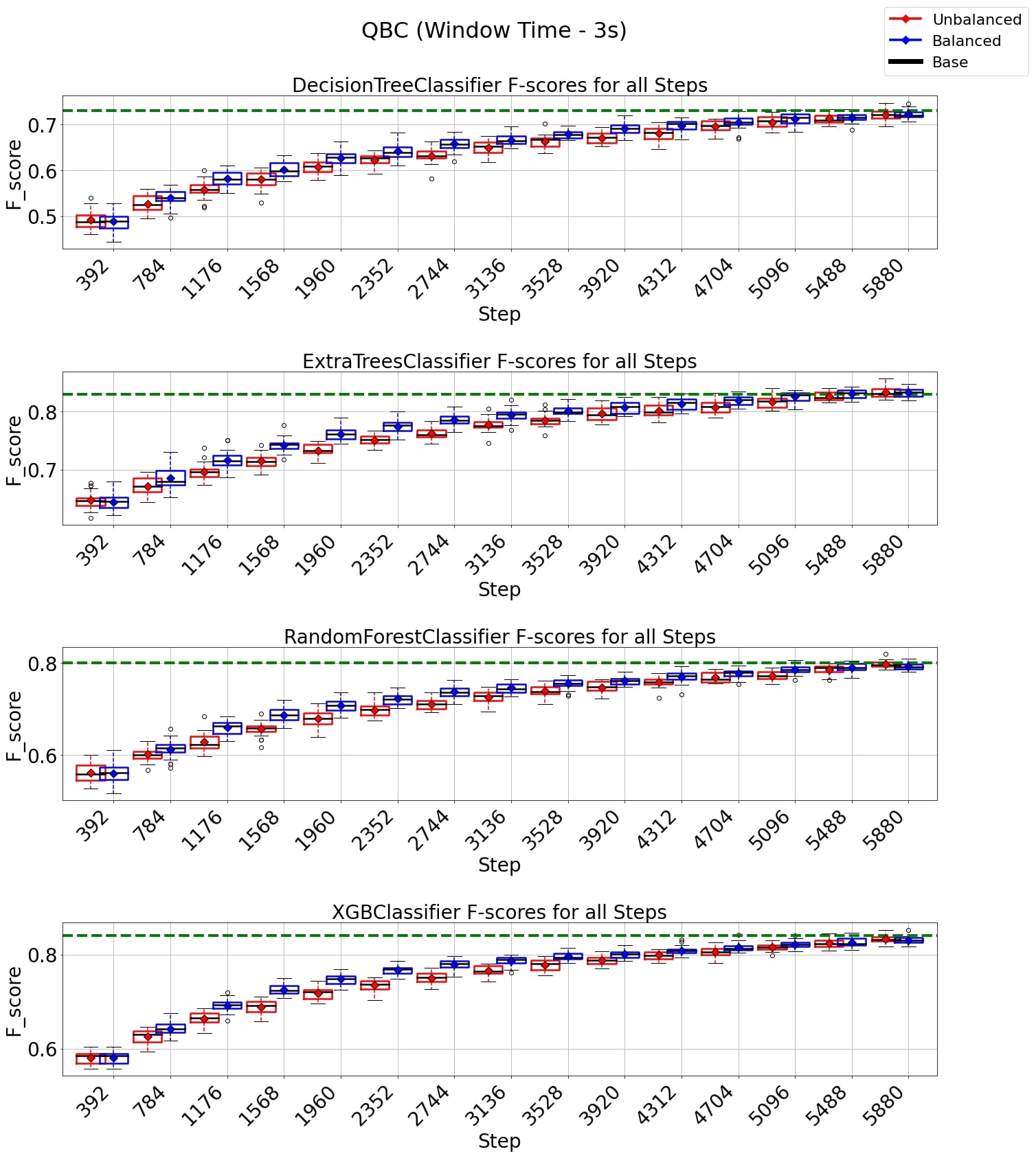}
    \caption{f1-score plots of the query by committee strategy for the 3-second window.}
    \label{qbc3s}
\end{figure}

\begin{figure}[!ht]
   \centering 
   \includegraphics[width=1\textwidth]{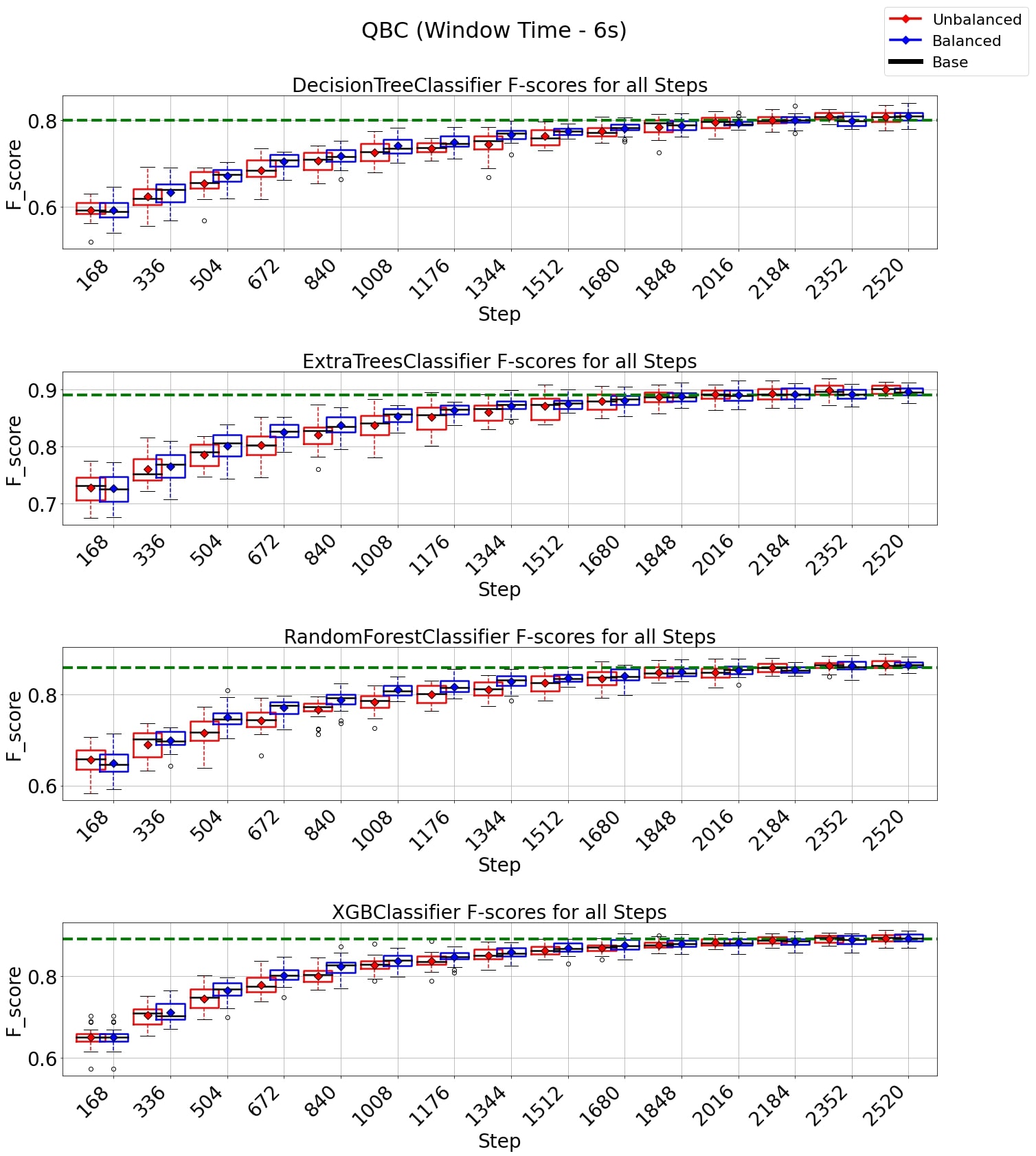}
    \caption{f1-score plots of the query by committee strategy for the 6-second window.}
    \label{qbc6s}
\end{figure}

In the case of the QBC strategy, all classification models reach at least their baseline score for a window of 3 seconds, as shown in Figure \ref{qbc3s} with a budget of 5880 instances from a total of 8399 instances (70\% of the entire data set). 
It is essential to observe from the experiments depicted in Figure \ref{qbc3s} that the class-balancing algorithm indeed increased the performance of the QBC strategy compared to the standard QBC implementation, mainly in the early cycles of the QBC strategy.  
For a window of 6 seconds (Figure \ref{qbc6s}), the QBC strategy reaches at least the baseline performance with 1512 instances (42\%, out of the 3,599 total) for all classification models using the class-balancing algorithm. 
This represents a reduction of 58\% in the training data used to achieve at least a similar performance. 
Moreover, a distinction is noticed between the QBC strategy with and without the balancement technique in its initial cycles, as the class-balancement algorithm shows considerable improvements compared to the standard QBC strategy at most steps evaluated in this experiment. 
Finally, it is essential to point out that using QBC, a window of 6 seconds, and the balancing strategy surpasses all baseline models' performances for this setup. 

\subsection{EMC Strategy on 3 and 6 seconds windows}

\begin{figure}[!ht]
   \centering 
   \includegraphics[width=.95\textwidth]{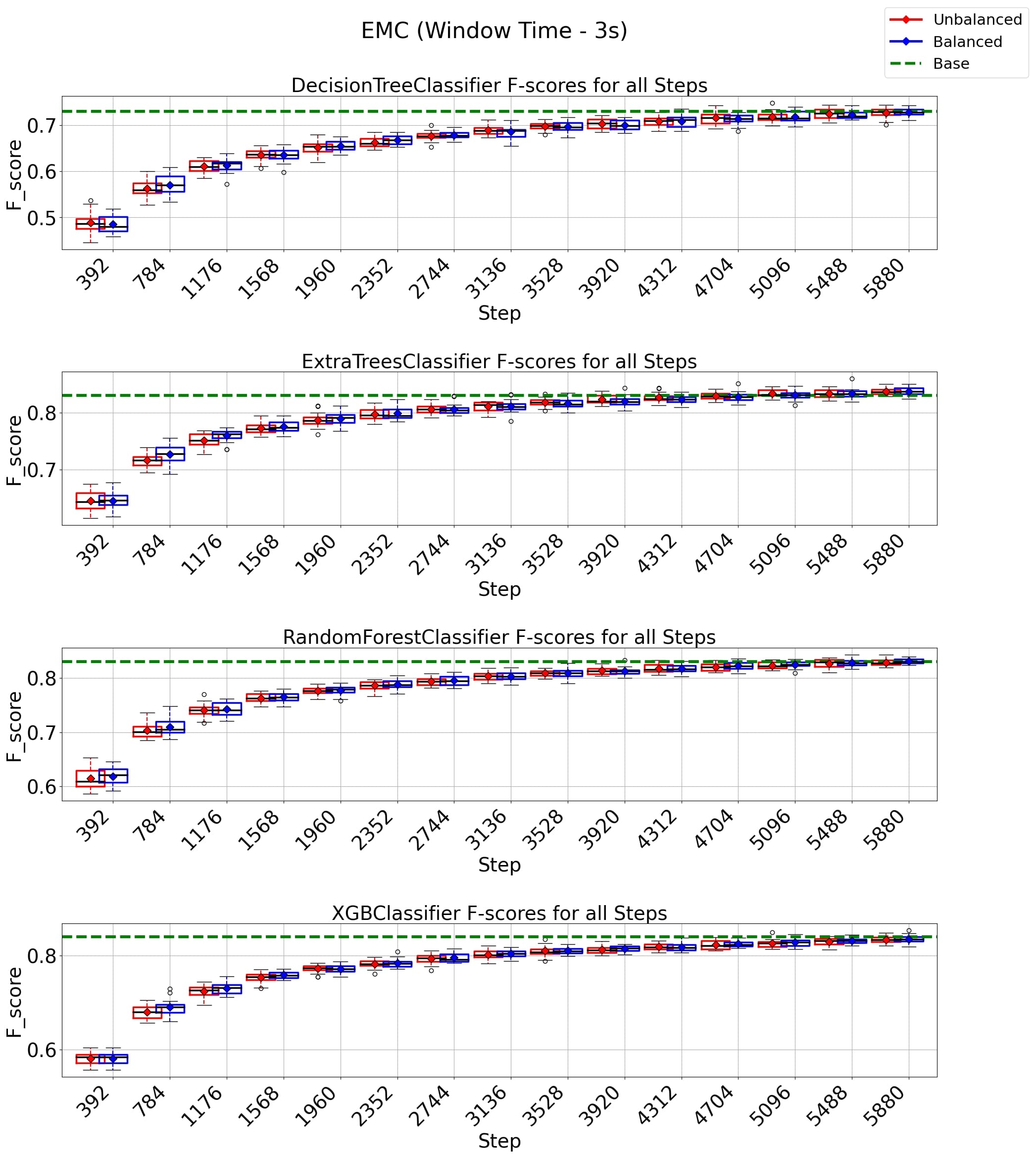}
    \caption{f1-score plots of the expected model change strategy for the 3-second window.}
    \label{emc3s}
\end{figure}

\begin{figure}[!ht]
   \centering 
   \includegraphics[width=0.9\textwidth]{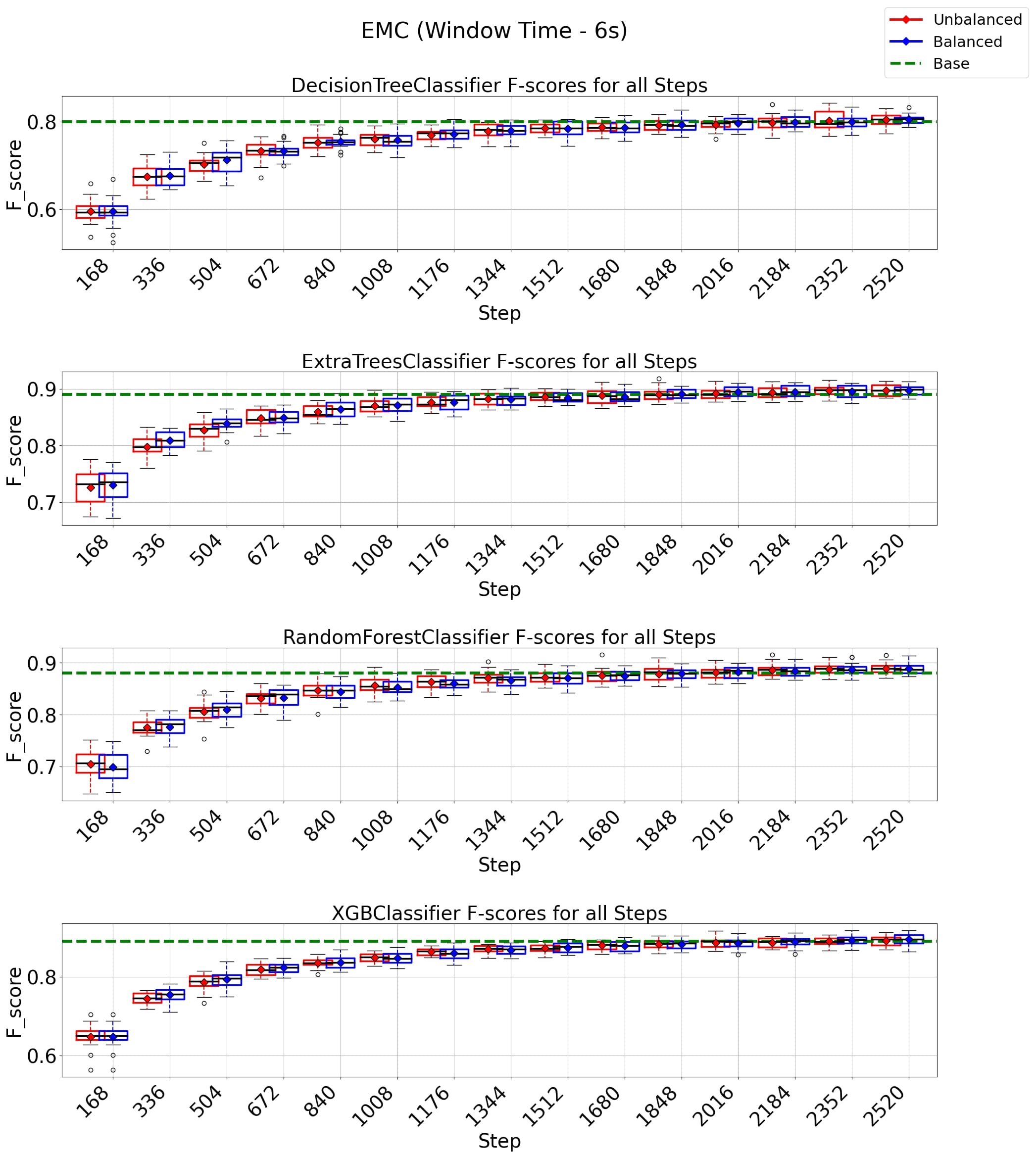}
    \caption{f1-score plots of the expected model change strategy for the 6-second window.}
    \label{emc6s}
\end{figure}

In the case of the EMC strategy, all classification models reach at least their baseline score for a window of 3 seconds, as shown in Figure \ref{emc3s} with a budget of 5880 instances from a total of 8399 instances (70\% of the entire data set). 
It is essential to observe from the experiments depicted in Figure \ref{emc3s} that the class-balancing algorithm indeed increased the performance of the EMC strategy compared to the standard EMC implementation, mainly in the early cycles of the EMC strategy.  
For a window of 6 seconds (Figure \ref{emc6s}), the EMC strategy reaches at least the baseline performance with 2,016 instances (56\%, out of the 3,599 total) for all classification models using the class-balancing algorithm. 
This represents a reduction of 44\% in the training data used to achieve at least a similar performance. 
Similarly to the QBC strategy, a distinction is noticed between the EMC strategy with and without the balancement technique in its initial cycles, as the class-balancement algorithm shows improvements compared to the standard EMC strategy at most steps evaluated in this experiment. 
Finally, it is essential to point out that using EMC, a window of 6 seconds, and the balancing strategy surpasses all baseline models' performances for this setup.

\subsection{Comparing Window Sizes and Active Learning Strategies Performances}

In this experiment, first, we evaluate the statistical significance of the difference between the window sizes (3 and 6 seconds) per model and AL strategy. 
Here, the objective is to ensure that the differences are statistically significant to affirm that one window size is better than the other for our problem. 
Therefore, we created Table \ref{Tab1} as follows. 
We used all experiments with the class-balanced strategy for both UNC and QBC, for all models and window sizes with the maximal budgets of 5880 (represents a total of 70\% of the processed data using our pipeline) for 3 seconds windows and 2520 (represents a total of 70\% of the processed data using our pipeline) for 6 seconds windows. 
These budget values were chosen to ensure all combinations of AL strategies and machine-learning models could have enough data to at least reach the baseline performance and, therefore, have a more fair comparison between all combinations.
We averaged all these experiments and also reported the standard deviations. 

As seen in Table \ref{Tab1}, using a window of 6 seconds always surpasses a window of 3 seconds regardless of the machine learning models and AL strategy adopted, with average f1-score differences at most 8\%. 
We conducted a Wilcoxon signed rank test between each model's average for a 3 and 6-second window and per AL strategy first. 
We verified that all these averages are indeed statistically significantly different for the window sizes of 3 and 6 seconds per AL strategy as the p-values were all below the threshold (i.e., with $p-value < 0.05$, we reject the null hypothesis that those averages come from the same distribution). 
Therefore, we can affirm that using a window size of 6 seconds is the best choice for our problem for all AL strategies.

Using Table \ref{Tab1}, we also verified which combination of a machine learning model, 6-second window, and AL strategy (using the balanced strategy) would perform best in our problem. 
As seen in Table \ref{Tab1}, an Extra Tree trained with Uncertainty Sampling and a 6-second window reached a 90.25\% f1-score, and therefore, it is the best result achieved. 
We executed a Wilcoxon signed rank again, comparing this best average value with all other combinations of the 6-second window and AL strategy (using the balancing strategy), and the statistical values showed that this choice was better than all others. 
Therefore, we conclude that Extra Tree trained with Uncertainty Sampling and the class-balance technique are our problem's best machine-learning model choices for our tactile sensing classification problem for a total budget of 70\% of the training data given to our entire data pipeline.

\begin{table}[!ht]
\centering
\small
\begin{tabular}{@{}lcccccc@{}}
\toprule
\multirow{4}{*}{Models} & \multicolumn{2}{c}{UNC} & \multicolumn{2}{c}{QBC} & \multicolumn{2}{c}{EMC} \\ \cmidrule(l){2-7} 
 & \multicolumn{2}{c}{Window size} & \multicolumn{2}{c}{Window size} & \multicolumn{2}{c}{Window size} \\
 & 3 seconds & 6 seconds & 3 seconds & 6 seconds & 3 seconds & 6 seconds \\ \cmidrule(r){1-7}
Decision Tree & 72.60 $\pm$1.26 & 80.67 $\pm$1.39 & 72.94 $\pm$1.00 & 81.12 $\pm$1.34 & 72.81 $\pm$0.98 & 80.52 $\pm$1.14\\
Extra Tree & 84.24 $\pm$0.71 & \textbf{90.25} $\pm$\textbf{0.92} & 84.21 $\pm$0.58 & 90.13 $\pm$1.03 & 83.72 $\pm$0.78 & 89.71 $\pm$0.95 \\
Random Forest & 83.53 $\pm$0.79 & 89.13 $\pm$1.24 & 83.44 $\pm$0.66 & 89.15 $\pm$1.09 & 83.10 $\pm$0.48 & 88.86 $\pm$1.13\\
XGBoost & 83.76 $\pm$0.86 & 89.80 $\pm$1.08 & 84.02 $\pm$0.87 & 89.58 $\pm$1.46 & 83.52 $\pm$0.81 & 89.48 $\pm$1.41 \\ \bottomrule
\end{tabular}
\captionsetup{justification=centering}
\caption{f1-score of AL strategies for different windows and models.}
\label{Tab1}
\end{table}

\section{Discussion}

In this paper, we tested and improved three standard AL strategies for a texture classification task. 
To the best of our knowledge, this is the first work on using AL for tactile texture classification.  
Although we used the same sensor and experimental process as in \cite{lima2021classification}, we worked on some limitations of their methodology. 
The authors in \cite{lima2021classification} had to undergo an expensive training process as they used 12 seconds of data exploration and trained their model on such a massive amount of data. 
Despite this significant effort, the training approach seemed to lack attention to the temporal features that could play a crucial role in the final outcome as they used each value from the sensor as a feature. 

Our work advances existing texture classification methodologies by proposing a novel pipeline for tactile data used for texture classification. This pipeline includes a time window approach for extracting features and using AL with a class balancing algorithm. 
Therefore, we addressed a significant issue in tactile data for texture classification: training a high-performance machine learning model with a limited number of training data.
Building such a high-performance machine learning model has several implications, including the possibility of deployment of low-complexity models in low-cost robotic hardware and the time reduction for the robotic arm exploration for classifying textures using tactile sensors.
Our proposed pipeline achieved competitive and even better classification performance than an established baseline, with less annotated data, making the learning process faster and more efficient. 
We show that using at most 70\% of the data available is enough to achieve and surpass the baseline performances using our proposed pipeline and algorithms.
Moreover, we considered the effect of temporal features using our sliding overlapping windows and extracting the sensors' data distributions. 
This gave the machine learning models detailed information regarding the robotic exploration, improving the performance of machine learning models learned from the tactile data. 
The AL strategies applied to the processed data further reduced the need for labeling training instances, as they selected only the most valuable ones to be labeled and given to the classifiers. 
As observed in the plots for 6-second windows, the learning process achieved the baseline results faster (with a lower budget for instances selected) and more efficiently (achieving an average f1-score of $90.21\%$) compared to 3-second windows.
We believe that such a result was because a window of 6 seconds in the conducted experiments ensures that the robot explores the texture in the two axes. 
Therefore, changes in texture in both axes could be better characterized by the extracted features. 
The results also show that in some of our 3-second-window experiments, the baseline method outperforms specific active learning configurations, specifically the Decision Tree with uncertainty sampling and all Query by Committee (QBC) combinations. 
We believe this can be attributed to several key factors.
Firstly, the discrepancy in performance between the baseline and active learning for the 3-second window experiments can be linked to the unique characteristics of our dataset and the duration of tactile exploration. 
Our dataset is tailored to focus on textures that primarily exhibit variations along the x and y axes. 
A 3-second window, however, typically encompasses only one dimension of exploration, resulting in limited diversity in the captured data. 
Consequently, when the entire labeled dataset is available for model training, simpler models like the Decision Tree may not perform better due to the absence of all data points. 
To optimize labeling efforts, we imposed a maximal budget of 70\% for active learning, beyond which further labeling would yield diminishing returns. 
In contrast, the more complex models employed in active learning demonstrate superior performance with a lower labeling budget. 
Moreover, the specific approach of bagging with randomly selected elements employed in QBC for creating the voting committee may inadvertently limit diversity in the committee models, affecting the method's performance in 3-second-window experiments.
In contrast, our 6-second window experiments demonstrate a different pattern, with active learning methods consistently outperforming the baseline. 
This can be attributed to the extended exploration duration covering at least two texture dimensions. 
This aligns more closely with the inherent characteristics of our dataset and showcases the efficacy of active learning strategies in scenarios where exploration durations offer a more comprehensive view of the tactile data's complexity.

Furthermore, our balancement algorithm makes our AL strategies robust to the distribution imbalance from queries performed by standard AL strategies. 
The balancing algorithm showed a marginal effect on the UNC strategy, but it improved the results for the QBC and EMC strategies, mainly in the early stages of our process.
Finally, combining a standard UNC strategy with the balancing algorithm and an ExtraTree as the machine learning model produced the best result for our tactile texture classification dataset. 
In conclusion, this work made the process of texture classification using tactile sensors more precise and efficient for real-world unstructured and dynamic environments by surpassing the results obtained by previous works.

We intend to expand this work in several directions. 
First, we would like to test and adapt our proposed pipeline for different types of tactile sensors and other sensory data. 
This approach would involve understanding how it performs with varying factors associated with different sensors. 
Extending this work to other dynamic real-world environments, such as industrial ones, would provide insights into the robustness of the pipeline. 
This work mainly focused on UNC, QBC, and EMC as active learning strategies.
Future research endeavors include exploring alternative Active Learning (AL) strategies to assess their potential advantages and insights within tactile texture classification. 
Another path we intend to explore is how AL would perform in scenarios with incomplete data, such as those caused by partial contact between the tactile sensor and the examined object. 
We believe AL has the potential to select those instances that a model deems more uncertain and, therefore, learn from them and adapt over time, gradually becoming proficient in handling such scenarios.

\chapter{Unbalanced Fault Classification using Active Learning in Synthetic Fiber Manufacturing Process}

\section{Introduction}

Automation, data analytics, machine learning, and artificial intelligence breakthroughs are driving a significant upheaval in manufacturing.
Recently, industrial processes have become more intelligent and data-driven, strongly emphasizing quality control and production efficiency. 
Such systems rely upon real-time monitoring of industrial processes and products. 
The standard method utilizes industrial sensors within the production process, e.g., \cite{meister2021review}. 
These sensors generate large volumes of time-series data, which must be processed and interpreted. 

Synthetic fiber is manufactured using polymers obtained from petroleum or natural gas. 
In principle, the process involves extruding a liquified polymer through small openings and solidifying it into fiber filaments through cooling, followed by complex post-processing to impart the properties downstream customers require. 
Moreover, several variations and customizations to the generic process outlined above exist, depending on the manufactured product. 
For producers in this industry, ensuring product quality, eliminating faulty fiber, and reducing production downtime are essential. 
Fiber defects in the manufactured product can result in costly claims from downstream customers. 
Therefore, reducing customer claims becomes critical to long-term success in the industry. 
 
To assist fiber manufacturers in achieving these aims, Instrumar Limited has developed a polymer fiber monitoring system \cite{chan2000new}, currently known as the Instrumar Fiber System (IFS). 
IFS uses electromagnetic sensors developed by Instrumar Limited to monitor the properties of the polymer fiber as it is being produced. 
Each sensor measures the physical properties of the fiber being produced and generates a stream of time-series data. 
These data are then analyzed to detect patterns corresponding to the fiber's or production processes' physical faults. 
IFS currently relies on manually identifying data patterns corresponding to physical problems and looking for such patterns in the data. 
As this process is labor intensive and susceptible to inaccuracies, an automated data analysis and fault detection process will help increase efficiency and reduce costs.
 
Detection of industrial production faults and manufactured product defects using machine learning and artificial intelligence based classification techniques have been used and proposed in \cite{wen2017new,de2015data,yu2019global,angelopoulos2019tackling,Rocha2020Dynamic, 
lima2021classification}. 
However, identifying and classifying faults in time-series data from sensors monitoring synthetic fiber production poses several unique challenges. 
Firstly, most of the data represents the normal functioning of the production process. 
Only a tiny fraction of the data represents defective products, which must be identified. 
Secondly, there are several product defects, each with its own pattern. Moreover, unidentified and anomalous data patterns often do not cause product defects and are not of interest to the customer. 
Consequently, labeling these datasets is labor intensive, requiring a substantial time commitment from domain experts, and hence costly. 
Finally, due to the variety of defect types with varying frequencies, labeled datasets will likely suffer from severe class imbalance. 

In this work, we investigate using AL \cite{settles2009active} strategies for fault classification. 
AL is based on the assumption that a machine-learning (ML) algorithm could perform better using a smaller number of labeled instances if we carefully choose only the most informative ones for the model to learn from. 
This reduces the need for a large volume of labeled data and hence saves data annotation time for industries, thereby reducing costs and enhancing efficiency.

The AL techniques we implement are integrated into the specific dynamics of synthetic fiber production and the various fault types that arise in its production process. 
We use time-series data from Instrumar sensors installed at a fiber manufacturing plant operated by one of their customers.
These data serve as a valuable resource for analyzing patterns and classifying the diverse data patterns representing faults of significant concern to the manufacturer.
Our goal is to enable effective classification of fault situations, minimize production disruptions, and ultimately improve fiber products' quality and reliability by utilizing AL's power. This research proposes an integrated approach, employing Active Learning techniques and a specialized class-balancing instance selection algorithm, to significantly reduce labeled data requirements, enhance accuracy, and effectively address class imbalance issues in fault classification for synthetic fiber manufacturing.
Through this study, we hope to positively impact the Synthetic Fiber Industry's continuous transition to intelligent manufacturing and quality control procedures.
This project aligns with the Fourth Industrial Revolution \cite{angelopoulos2019tackling} objective of developing data-driven, intelligent manufacturing processes. 

The present paper provides a literature review in Section~\ref{lit}. 
Our methodology and pipeline are presented in Section~\ref{met} followed by results and discussion in Sections~\ref{res} and \ref{dis}, respectively.


\section{Literature Review \label{lit}}


Rapid industrial evolution in fiber manufacturing necessitates increasingly accurate and effective fault classification techniques.
Over the past few decades, model-based approaches and data-driven methodologies have made substantial progress in fault diagnostics. 
With the ever-increasing use of sensors and IOT in the production process, data-driven approaches have gained popularity \cite{peng2020cost}. 
This literature review delves into data-driven methodologies that promise to transform fault classification in industrial production.

The authors of \cite{caiazzo2022towards} have identified data-driven approaches as supervised and unsupervised, with the former requiring labeled training datasets while the latter does not. 
Regression and classification are the most commonly used ML tasks in manufacturing, and supervised learning tends to be the dominant approach, with unsupervised learning being far less used \cite{kang2020machine, de2015data}. 
Moreover, abundant production line data has also led industries to focus more on supervised techniques. Angelopoulos et al. \cite{angelopoulos2019tackling} highlighted the necessity of vast amounts of data for machine learning algorithms to be adequately trained for use in many industrial decision-making processes. 
Fernandes et al. \cite{fernandes2022machine} point out that Decision Trees tend to be suitable for applications in an industrial context. 
In \cite{amihai2018industrial,binding2019machine}, Random Forest models were used because of the efficiency and effectiveness of ensemble tree models. 
Sacco et al. \cite{sacco2020machine} presented a cutting-edge inspection method for  Automated Fiber Placement (AFP) that uses a deep convolutional neural network to classify defects on a per-pixel basis. However, due to the need for large, clearly annotated training datasets, these supervised learning techniques may not be suitable for application in industrial projects using AFP  \cite{ghamisi2023anomaly}. 
According to \cite{heinecke2019manufacturing}, the lack of labeled training data is primarily caused by the high cost and inconvenience of gathering real-world data from production machines, the rarity of defects that necessitate extensive data collection, and the absence of standardized guidelines for human inspectors to identify, document, and label anomalies and defects for machine learning. 
It is difficult to translate the multiplicity of standards and practices into a clear labeling strategy for training models \cite{ghamisi2023anomaly}. To detect and classify faults, supervised models need to be trained on faulty instances, which are rare in manufacturing. To address this issue, research has been done on creating synthetic datasets in \cite{sun2021quality} for detecting and classifying the wielding of carbon fiber and in \cite{meister2021synthetic} to detect defects in AFP. According to the authors of \cite{ghamisi2023anomaly}, although such techniques of data generation have shown effectiveness, they still need suitably large and representative datasets. Furthermore, it takes a lot of computational power for industries to thoroughly process enormous volumes of data to arrive at a solution that is close to ideal \cite{angelopoulos2019tackling}. Because of this, supervised learning models are infeasible in actual industrial settings.

Although labeled data is scarce, there is still a lot of unlabeled data available in the industry. To utilize these unlabeled data, research has been conducted to incorporate labeled and unlabeled data to train the models, a technique known as semi-supervised learning \cite{fredriksson2022machine}. 
In \cite{li2021novel,wu2021hybrid}, the authors studied the effect of semi-supervised methods for fault diagnosis. 
Although these semi-supervised methods yield good performance, significant limitations are associated with them. 
One of them 
is the complexity of the training process, which results in increased training time and a higher requirement for computational resources. 
Besides, it remains unclear how well the results, such as those of \cite{li2021new}, generalize to other types of fault signals.

\begin{figure}[!htbp]
  \centering
  \begin{minipage}[t]{0.3\textwidth}
    \centering
    \includegraphics[width=\linewidth]{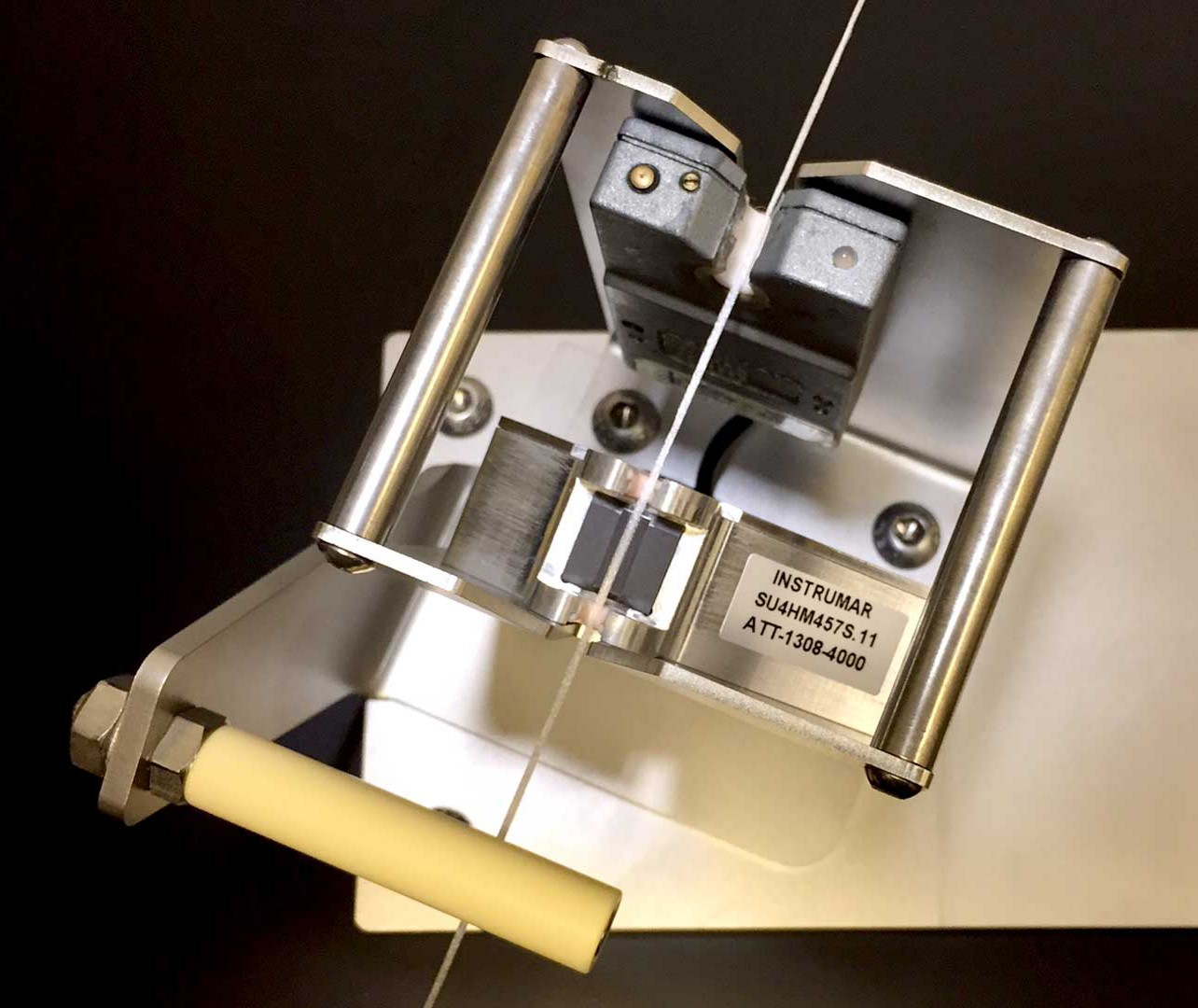}
    \caption{Instrumar's fiber quality monitoring sensor.}
    \label{sensor}
  \end{minipage}%
  \hfill
  \begin{minipage}[t]{0.3\textwidth}
    \centering
    \includegraphics[width=\linewidth]{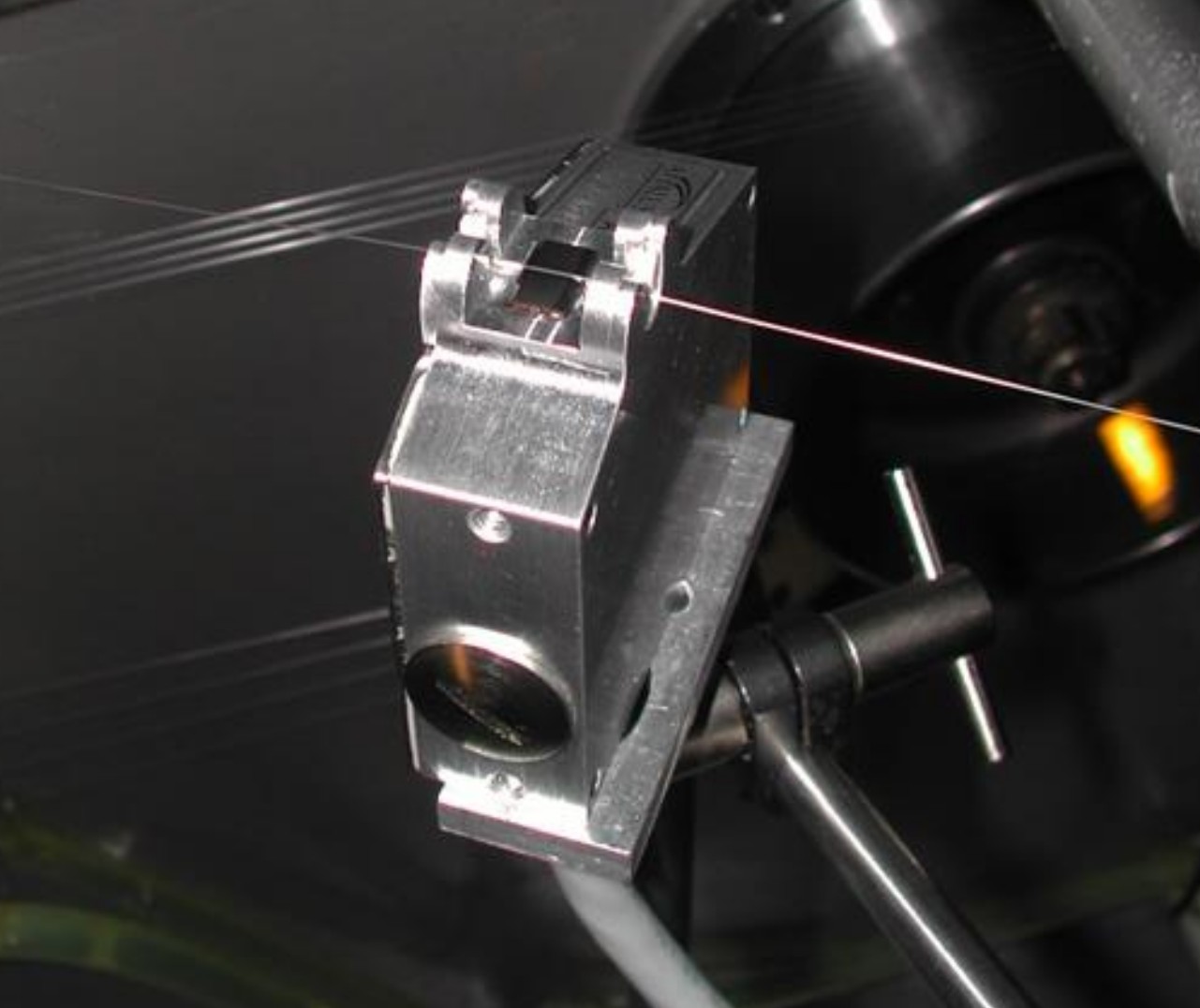}
     \caption{Instrumar's fiber quality monitoring sensor at a production plant.}
      \label{second_sensor}

  \end{minipage}%
  \hfill
  \begin{minipage}[t]{0.3\textwidth}
    \centering
    \includegraphics[width=1.2\linewidth, height=1.5in]{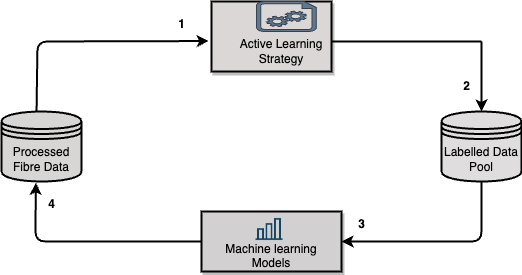} 
    \caption{Active Learning pipeline for fault classification.}
    \label{pipeline}

  \end{minipage}
\end{figure}

Problems with data-based modelling are frequently encountered because the training process is hampered by data loss, redundancy, mislabeling and class imbalance
That is why the authors have prioritized improving the efficiency and processing time of learning algorithms and creating learning strategies that can successfully adjust to different levels of uncertainty. 
To achieve such control, AL can be a solution that merits further exploration by industry. The use of AL substantially reduces the need for labeled data by allowing the model to actively choose the most informative data points to be labeled (i.e., annotation) by an expert. 
The core concept of AL is that if the learning algorithm can choose the data from which it learns, it will perform better with less training data \cite{settles2009active}. 
AL techniques have employed many different query strategies, which are ways of choosing the most informative data points for labeling, with Uncertainty (UNC) sampling and Query By Committee (QBC) \cite{settles2009active} being the most commonly used. 
In \cite{peng2020cost}, the authors develop a cost-sensitive AL framework that 
reduces data wastage.  
Additionally, the authors attempted to tackle the imbalance between normal and fault-type samples. The authors in \cite{del2021active} have also used AL to classify industrial process time series data consisting of vibration waveforms or process control data. 
Authors in \cite{wu2021rlad} obtained sufficient annotated training data for multivariate time series classification using AL.

Although the imbalance between normal and defective samples in manufacturing datasets has been addressed in the literature, the imbalance among different classes of defective samples must be explored. 
This study investigates using AL strategies on real-life fiber industry time-series data to classify different types of faults in the production process under the constraint of a substantial imbalance between the numbers of different fault types. 

Our research introduces a pioneering approach to fault classification within industrial fiber manufacturing by devising a novel class-balancing instance selection technique integrated with Active Learning strategies. This innovation stands as a significant progression beyond existing studies, which predominantly focus on Active Learning in fault classification. Our method addresses the crucial industry challenge of class imbalance among production fault types, which is a  critical aspect for fault classification methodologies, thus enhancing fault classification accuracy while notably reducing labeled data requirements.

\section{Methodology\label{met}}

In this section, we present our comprehensive framework for enhancing fault classification within the domain of industrial fiber manufacturing.
This methodology begins with a detailed exploration of data collection, followed by preprocessing and feature extraction techniques, ensuring data quality and readiness for analysis.
Here, we introduce AL strategies, showcasing their potential in optimizing the fault classification process.
We also present our novel class-balancing instance selection algorithm that aims to mitigate the impact of data imbalance on AL strategies and improve classification accuracy. 
Finally, we discuss the selected classification models and relevant metrics for evaluation.

\subsection {Data Collection, Preprocessing and Feature Extraction}

Here, we delve into the critical foundation of our study, where we leverage real-world factory data provided by our industry partner, Instrumar Limited. 
The time series dataset utilized in this work originates from Instrumar's proprietary sensors, shown in Figure \ref{sensor}, specifically designed to monitor industrial fiber manufacturing processes.

The Instrumar Fiber Sensor Unit (SU), installed on a production line, measures the electrical impedance of fiber as it passes through an electromagnetic field. 
In effect, the sensor is sensitive to the electrical properties and the geometry of the fiber. 
The electrical impedance response signal is measured at 20 kHz to 40 kHz. 
These data are then processed through Instrumar's Sensor Processing Unit (SPU), which calculates and outputs four fiber properties every 200ms: (i) $magnitude$, (ii) $phase$, (iii) $node\ quality$, and (iv) $node\ count$. 
The $magnitude$ is related to the $denier$ or the density of the fiber. 
Nodes in the fiber cause a drop in the electrical response signal. 
The frequency of such drops is the $node\ count$, while the amplitude of the drops is the $node\ quality$. 
The $phase$ measures the time delay in the response signal. 
It is sensitive to conductivity and hence the amount of finish applied to the fiber. 

The four fiber properties described above are measured every 200ms and are then used to detect defects in the fiber and infer the underlying process issues that could have produced such defects. 
Besides serving as the input for Instrumar's current system, they represent a valuable resource that enables us to explore and implement our proposed research for fault classification in this industrial context.
 
 Instrumar's current data analysis system matches the data to known patterns corresponding to fiber defects or physical events. 
 Instrumar's engineers identify these data patterns and build heuristic algorithms to catch them, known as $custom\ fault\ alarms$. 
 This process tends to be labor-intensive and error-prone. 
 Additionally, setting the parameters for these algorithms is challenging. 
 They often need to be fine-tuned and adjusted to minimize false positives or capture missed events based on customer feedback. 
 Moreover, they are blind to large data fluctuations with a previously unknown or unidentified pattern. 
 This could potentially result in products that do not meet specifications being passed off as normal, leading to costly customer claims and substantial financial loss.   
 
Instrumar aims to replace this system with a more robust two-stage process. 
In the first stage, data are processed through an unsupervised learning algorithm to identify abnormal or anomalous data. 
These anomalous data are then classified by a multiclass classification algorithm into specific faults, which are then communicated to the customer to initiate further feedback actions. 
Neither of these algorithms can be run on the raw time-series data. 
Instead, the time-series data must be further processed. 
The data are divided into time windows, and a set of features is extracted for each window for each fiber property. 
The set of features consists of the mean, median, standard deviation, variance, chi-square, 25th percentile, 75th percentile, minimum and maximum values, kurtosis, skewness, and stability (i.e., the ratio of current fiber property level to average level over last 24 hours) totalling 12 features that are extracted for each of the four fiber properties. 
Thus, the total number of features extracted is 48.  
Each window with its set of features constitutes a data instance. 
 
Our goal in this study is to examine and evaluate the use of AL techniques to classify anomalous fiber property data into various fault types. 
For the training and evaluating our models, we use the labels provided by the $custom\ fault\ alarms$. 
While these labels are not necessarily perfectly valid, they serve as a valuable control sample to evaluate and validate our results against and inform Instrumar's business decisions about the feasibility of deploying the new fault identification system. This work uses four different $custom\ faults$, namely $crossover$, $entanglement$, $godetwrap$, and $major\ crossover$.
The $crossover$ is a condition where a small number of filaments from one fiber bundle migrate to another fiber bundle. An $entanglement$ is a limit-based alarm that fires if the number of knots expected deviates past a minimum and maximum limit. 
A $godetwrap$ event is when several filaments break and wrap around one part of the machine, the godet. A $major\ crossover$ is like $crossover$ except that it causes the line to break very quickly, so a large migration of filaments causes significant issues.



\subsection {Data Class Imbalance}

The distribution of fault types in our dataset shows a substantial disparity, with specific fault categories  
being far more common than others. The number of instances for $crossover$ is 2485, $entanglement$ is 35, $godetwrap$ is 860 and $major\ crossover$ has 221.
That is, the number of $crossovers$ is 70 times higher than the number of $entanglement$ events in our sample. 
This class imbalance arises because of significant differences in the frequencies of different types of faults in the manufacturing process.


\subsection {Active Learning Strategies and Class-Balancement Instance Selection Algorithm} 

AL techniques aim to choose the most uncertain examples the model has identified and have them labeled by an annotator.
Our study compares Uncertainty sampling (UNC) \cite{settles2009active} and  Query by Committee (QBC) with bagging \cite{settles2009active} for the purpose of fault detection in the fiber industry.
The UNC approach iteratively chooses from a pool of unlabeled data the instances that will be most useful for labeling.
Using metrics like entropy, margin sampling, or least confident predictions, it assesses how unsure the model is about its predictions.
The model is then updated during training using the labeled data.
Another AL technique, QBC with Bagging, combines QBC with a bagging ensemble. 
Employing bootstrapped sampling to produce various models also detects unclear labels by analyzing model committee disagreement.
The examples with a high level of uncertainty are labeled, and as new labeled data is added, the ensemble is updated.



Figure \ref{pipeline} shows the pipeline developed for this work.
In our work, we split the processed data into two sets: a training set with 80\% of the data and a test set with 20\% of the data for validation. 
The processed fiber data in Figure \ref{pipeline} serves as the unlabelled pool in our AL method from which the most ambiguous labeled instances are queried for labeling by the AL strategies (step 1).
We do not use a human annotator to annotate the data in our tests because the labels are retrieved from the Instrumar $custom\ fault$ alarms.
Once an instance has been labeled, it is added to the collection of labeled data, as in step 2. 
Then, we train a machine learning model using the instances in the labeled pool (step 3) and use the trained model to classify the instances present in the processed pool of fiber data (step 4). 
Given that AL is an iterative process, we establish a maximum annotation budget, i.e., the maximum number of instances that can be queried throughout the process and a step size that specifies the number of instances from the unlabeled pool to be labeled at each AL iteration. 
Following these steps, steps 1 to 4 are iteratively executed until the annotation budget is reached. 
To ensure experimental reproducibility, we employed 20 seed values.
Our seed generation technique uses the successive decimal digits of $\pi$ in sets of four (seed1 = 1415, seed2 = 9265, etc.). 
Such a choice prevents the selection of seeds at random.
Instances are randomly queried during the first iteration of the AL techniques (step 1 in Figure \ref{pipeline}). 
For UNC, we employ the Least Confidence metric to assess the degree of uncertainty of instances in Equation 1, where  $y^*$ is the class label that a machine learning model most likely assigned.

\begin{equation}
    \phi^{\text{LC}}(x) = 1 - P(y^* | x; \theta)
\end{equation}

We have developed a committee with basic models trained on various subsets of labeled data while performing QBC with bagging. 
The committee of models collectively predicts the labels for unlabeled instances.
The level of disagreement among the committee members is assessed to determine the uncertainty or informativeness of each candidate sample, and each model in the committee offers its forecast for each occurrence.
To determine the degree of disagreement among the committee members in our instance, we employed the vote entropy, depicted in Equation 2 where $V(y_{t,m})$ is the number of votes a specific label m obtains from the committee of classifiers, C.

\begin{equation}
    \phi^{VE}(x) = -\frac{1}{T} \sum_{t=1}^{T} \sum_{m=1}^{M} \frac{V(y_{t,m})}{C} \frac{\log V(y_{t,m})}{C} 
\end{equation}

The class imbalance problem in the data is addressed by the novel class-balancing instance selection algorithm presented in this paper. 
The primary goal is to query the unlabeled instance pool so that the labeled data has a balanced representation of all classes. 
This strategy seeks to avoid bias in the model's classification performance, which could happen if some classes receive an excessively high number of instances from an AL strategy. 
Our algorithm mainly aims to modify the training set's class frequencies by inverting their occurrences. 
As a result, we reduce the number of instances picked from classes with more occurrences while prioritizing selecting more instances from classes with fewer instances in the training set. 
Integrating this algorithm with the AL strategies, we look to provide a balanced training set for our classifiers to train on. 
By aiming for a more fair representation of classes in the training data, our proposed class-balancing instance selection algorithm provides a viable technique to resolve a class imbalance in AL, limiting bias, which results in increased classification performance, as discussed in the Results section.
It is essential to understand that, given the class label of the chosen instance's uncertainties, which the expert will finally resolve, perfect balance cannot be guaranteed.
The pseudo algorithm for the balancement technique is described in Algorithm \ref{alg:compute-inverted-frequency-ch4}.

\begin{algorithm}
\caption{Computation of inverted example frequency.}
\label{alg:compute-inverted-frequency-ch4}
\hspace*{\algorithmicindent} 
\textbf{Input:} $current\_train\_y$: all label values used in the current training set\\
\hspace*{\algorithmicindent} \textbf{Output:} $inverted\_freq$: The inverted frequency of the class distribution 
\begin{algorithmic}[1]

\STATE $freq \gets \text{bincount}(current\_train\_y)$
\STATE $norm\_freq \gets \frac{freq}{\text{len}(current\_train\_y)}$
\STATE $sorted\_freq \gets \text{sort}(norm\_freq)[::-1]$
\STATE $sorted\_indices \gets \text{argsort}(\text{argsort}(norm\_freq))$
\FOR{$i$ \textbf{in} $sorted\_indices$}
    \STATE $inverted\_freq[i] \gets sorted\_freq[i]$
\ENDFOR
\RETURN $inverted\_freq$
\end{algorithmic}
\end{algorithm}

This procedure improves class distribution balance, which is essential for improved machine-learning algorithm performance.
The algorithm begins in line 1 by determining how frequently each class appears in the training set.
An array called $freq$ is produced by counting the instances of unique labels in the $current\_train\_y$ using the $bincount()$ function.
Line 2 normalizes the frequency values acquired in the first step to a range of 0 to 1.
A new array called $norm\_freq$ is produced by dividing each frequency value by the total number of examples in the training set (len($current\_train\_y$)).
The $norm\_freq$ array is sorted in descending order using the $sort()$ function, and the $[::-1]$ slicing is then used to reverse the order of the sorted array to produce an inverted probability distribution.
The array that is produced is called $sorted\_freq$ (line 3). 
Afterward, an auxiliary array called $sorted\_indices$ is created to identify the mapping that would cause the probabilities to be reversed (line 4).
When the $argsort()$ function is used twice to sort the $norm\_freq$ array, the indices that would result are stored in this array and would be sorted in ascending order.
After that, a loop that iterates over each element in the $sorted\_indices$ is run (lines 5 through 7). 
The corresponding value from the $sorted\_freq$ is assigned to the $inverted\_freq$ list at position $i$ for each index $i$. 
The $inverted\_freq$ values for every class in the training set are now included in the $inverted\_freq$ list, which the algorithm has created after all iterations and is returned as the algorithm's output in line 8. 

To summarize the algorithm, the technique reduces the prevalence of classes with higher representation while selectively adjusting the class frequencies to emphasize classes with lower representation.

\subsection {Classification Models and Evaluation Metric}

As AL is a strategy rather than a specific model by itself, we can integrate any classifier as a machine-learning model for the purpose of training and prediction. 
For our work, we have used the existing classifiers from the python scikit-learn package, such as XGBoost \cite{chen2016xgboost}, Random Forest \cite{breiman2001random}, and Gradient Boosting \cite{friedman2001greedy}. 
We have used the default version of these classifiers and have not performed any hyperparameter tuning. 
The reason for using the default version of these classifiers is to keep the models simple. Hyperparameter tuning can be considered as a further refinement if necessary.

Moreover, we have used the f1-score as our evaluation metric. Although accuracy is a commonly used metric for evaluating the performance of classification algorithms, it is not well suited for cases where the data suffer from class imbalance \cite{sokolova2009systematic}. 
In scenarios like ours, where the class distribution is heavily imbalanced, the f1-score considers precision and recall, making it a more informative evaluation metric for fault classification.

\section{Results\label{res}}

This section thoroughly examines the performance of our AL techniques, utilizing three different classifiers and the f1-score as the evaluation metric. 
We train and test each classifier using 5-fold cross-validation to determine their baseline f1-score performance.
Since the baseline scores indicate an average performance of the models using the complete dataset for model building, they can be utilized as a benchmark when evaluating the machine learning models and AL techniques employed in this work.
In this section, we also analyze the effect of our class-balancing instance selection on the AL strategies and evaluate them with and without the class-balancing algorithm. 
The description and analysis of these experiments are shown in Section \ref{sec:UNC} and Section \ref{sec:QBC}.
In Figure \ref{unc} and Figure \ref{qbc}, the horizontal axis (labeled Step) shows the number of instances queried to the annotator for labeling at each AL iteration, and the vertical axis shows the corresponding f1-score. 
For each step, we have two box plots. 
The red box plot represents the AL strategy's performance without the class balancing algorithm, and the blue box plot represents the Al strategy with the class balancing algorithm. 
In Section \ref{sec:750and2000}, we first compare the results of the AL strategies with the balancing technique using only 750 instances since we observed that all baselines were achieved at this step. 
Last, we compare and discuss the results of the AL strategies using the total budget of 2000. 
The main objective of Section \ref{sec:750and2000} is to determine if there is indeed a  combination of AL strategy and machine-learning model for this problem that achieves high performance with fewer instances. 

\subsection{Uncertainty Sampling Strategy}
\label{sec:UNC}

\begin{figure}[!ht]
   \centering 
   \includegraphics[width=1\textwidth]{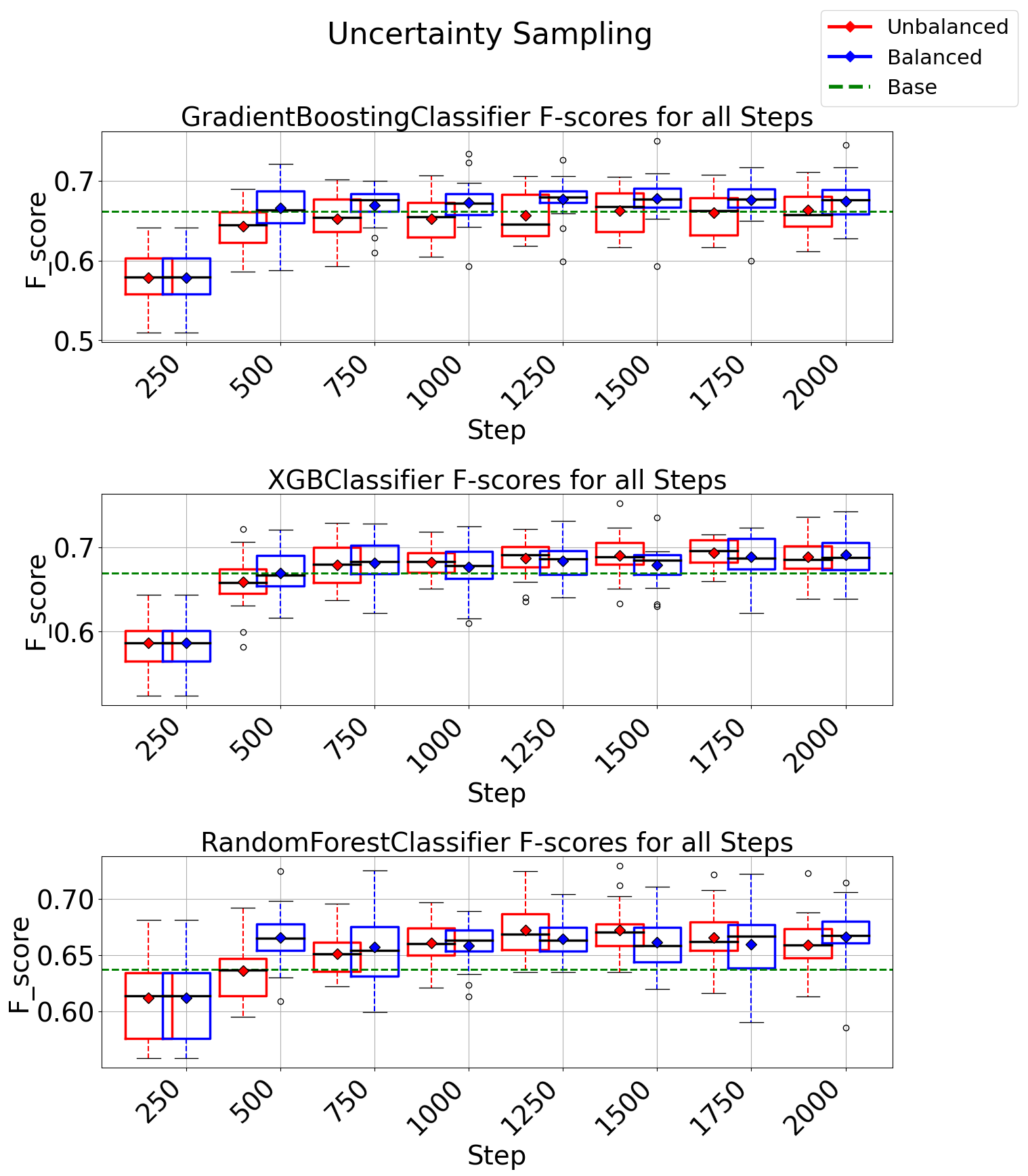}
    \caption{f1-score plots of the UNC strategy.}
    \label{unc}
\end{figure}

In our research, we implemented the Uncertainty Sampling (UNC) strategy on the processed anomaly data to reduce the number of instances labeled for a satisfactory fault classification performance.
In Figure \ref{unc}, we show the results of the UNC strategy. 
We noticed that within a fixed annotation budget, all classifiers using the UNC strategy were able to outperform their respective baseline f1-score values. 
The results suggest that the number of training instances for all the classifiers can be reduced to 750 out of a total of 3601 instances.
That is, the required number of training instances to surpass the baselines is only around 21\% of the total training data we have, and thus, the need for annotation can be reduced by 79\%.
It is also noticeable that UNC with the class-balancement algorithm performed considerably better than the one without the balanced strategy.
This effect is particularly prominent during the early stages of the AL process (i.e., using 500 to 1000 selected by the UNC strategy), where the benefits of class balancing become more noticeable.

\subsection{QBC Strategy}
\label{sec:QBC}

\begin{figure}[!ht]
   \centering 
   \includegraphics[width=1\textwidth]{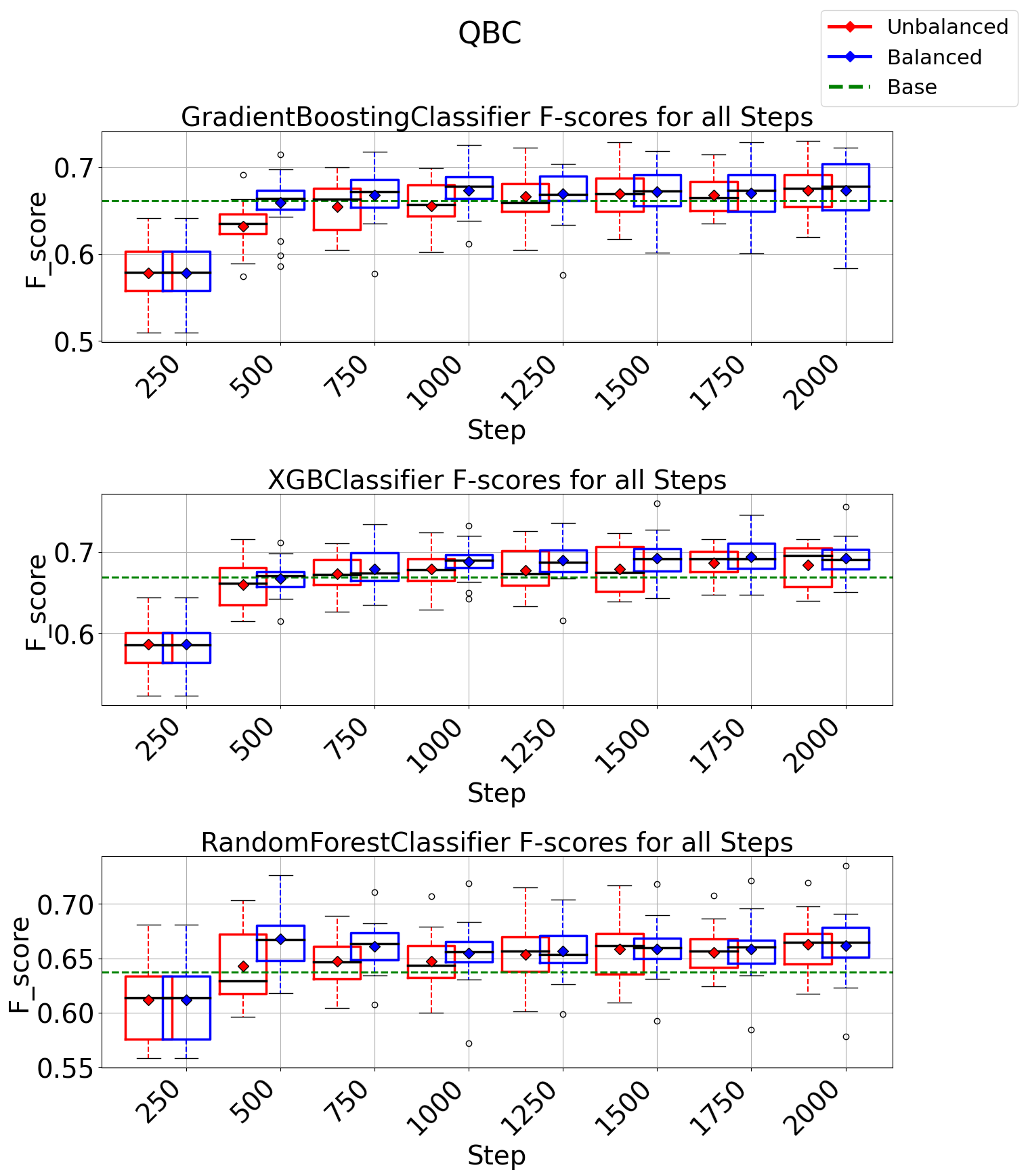}
    \caption{f1-score plots of the QBC strategy.}
    \label{qbc}
\end{figure}

For the QBC strategy, the results are illustrated in Figure \ref{qbc}. 
Notably, all classifiers using this strategy could outperform their respective baseline score with a budget of around 750 instances from a total of 3601 instances.
This also results in a 79\% reduction of the required labeled training set to classify faults. 
Moreover, we also see a difference in the performance of the class-imbalancement algorithm for different classifiers. 
Notably, for the Gradient Boosting classifier, QBC with the balancement algorithm performs better than QBC without the balancement technique at all steps. 
For XGBoost, the impact of the class-balancing technique is evident throughout all stages except the final step, whereas, for Random Forest, gains are observable by using the balancement algorithm in the early stages (i.e., 500, 750 and 1000 instances selected by QBC).

\subsection{Comparing AL strategies at steps 750 and 2000}
\label{sec:750and2000}

This section compares the performances of the two AL strategies UNC and QBC, per model, at steps 750 and 2000. 
For both AL strategies, we employ the class-balancing strategy in all of our experiments. 
The purpose of selecting these steps is to examine the performances of AL strategies when they outperform baseline results and after they have reached the annotation budget. 
Tables \ref{tab1} and \ref{tab2}, respectively, show the outcome of the comparisons for steps 750 and 2000. 
Each row represents a specific machine learning model, showcasing the respective average f1-scores and standard deviations achieved when employing both UNC and QBC strategies. 
The results in Tables \ref{tab1} and \ref{tab2} show that the machine-learning models perform nearly similarly for UNC and QBC. 

In Table \ref{tab1}, a statistical analysis was conducted to assess the significance of the average differences among various combinations of Active Learning (AL) strategies and Machine Learning (ML) models, employing the Wilcoxon signed-rank test. 
A p-value threshold of $0.05$ was utilized to determine statistical significance. 
Rejection of the null hypothesis that the averages originate from the same distribution implies a statistically significant difference.
The best outcome was observed with the Uncertainty Sampling (UNC) strategy combined with the XGBoost algorithm, yielding an average f1-score of 68.18\%. 
Therefore, we can conclude that with a budget of 750 instances, we could use any AL strategy with the XGBoost algorithm to achieve the best performance. 

Further examination of the entire budget of 2000 instances (refer to Table \ref{tab2}) revealed that the QBC and XGBoost combination outperformed all others, yielding an average f1-score of 69.24\%. 
A subsequent Wilcoxon signed rank test established its superiority over most configurations, though not reaching statistical significance compared to Uncertainty Sampling with the same ML algorithm.

In summary, the comprehensive analysis of Tables \ref{tab1} and \ref{tab2} leads to the conclusion that both AL strategies, UNC and QBC, effectively diminish the necessity for an extensive labeled dataset, achieving best performance in fault classification with the XGBoost algorithm using 2000 instances. 
While a considerable performance is attainable with a reduced budget of 750 instances, statistically significant enhancements are noticeable with the larger budget of 2000 instances.


\begin{table}[!ht]
\centering
\caption{Comparison of machine learning models' performances using AL strategies and the balancing algorithm at step 750.}
\begin{tabular}{@{}lcccc@{}}
\toprule
\textbf{}         & \multicolumn{2}{c}{\textbf{UNC}} & \multicolumn{2}{c}{\textbf{QBC}} \\ 
\textbf{Model}    & \textbf{f1-score} & \textbf{std.} & \textbf{f1-score} & \textbf{std.} \\ \midrule
Gradient Boosting & 66.97               & 2.35           & 66.78               & 3.00           \\
Random Forest     & 65.72               & 3.27           & 66.08               & 2.24           \\
XGBoost           & \textbf{68.18}               & \textbf{2.75}           & \textbf{67.92}               & \textbf{2.40}          
\label{tab1}
\end{tabular}
\end{table}

\begin{table}[!ht]
\centering
\caption{Comparison of machine learning models' performances using AL strategies and the balancing algorithm with the total budget of 2000.}
\begin{tabular}{@{}lcccc@{}}
\toprule
\textbf{}         & \multicolumn{2}{c}{\textbf{UNC}} & \multicolumn{2}{c}{\textbf{QBC}} \\ 
\textbf{Model}    & \textbf{f1-score} & \textbf{std.} & \textbf{f1-score} & \textbf{std.} \\ \midrule
Gradient Boosting & 67.47              & 2.66           & 67.35               & 3.67           \\
Random Forest     & 66.63               & 2.70           & 66.13               & 3.15           \\
XGBoost           & \textbf{69.08}               &   \textbf{2.57}           &  \textbf{69.24}               &  \textbf{2.30}          
\label{tab2}
\end{tabular}
\end{table}

\section{Discussion\label{dis}}

Our research shows the potential of AL techniques applied to real-world problems relevant to industry, such as classifying faults in fiber manufacturing using Instrumar's fiber monitoring data. 
The central insight from our findings is that AL, in conjunction with our class-balancing technique, provides a measurable and valuable advantage over conventional supervised machine learning techniques for fault classification, mainly by decreasing the need of a vast number of instances from which machine learning models can be trained.

Application of these techniques will likely help increase efficiency and reduce costs incurred due to fiber manufacturing faults. 
Our work emphasizes tackling the inherent imbalance between different types of faults in datasets related to industrial fiber manufacture.
Conventional machine learning approaches face significant challenges due to the magnitude of the required labeling effort and the strong class imbalance in the labeled datasets.
However, AL can tackle such real-life scenarios by judiciously selecting data points for labeling, especially with the incorporation of our class-balancing technique.
It made it possible for the classifiers to adjust and perform exceptionally well when categorizing even the rarest error types. 
Our study's optimization of the fault classification procedure while minimizing processing and annotation expenses is a noteworthy accomplishment. 
Through a proactive search for useful samples and class balance, our suggested methods improved efficiency and helped find faults at a reasonable cost. 
The AL process achieved the baseline results with five times fewer labeled instances, resulting in substantial gains in training efficiency and optimization.
This result has enormous practical implications for sectors such as the fiber manufacturing industry, where operating costs are a significant concern.
Our study presents a potential method that emphasizes data-driven and effective production processes in line with the Industry 4.0 goal.
We have shown how to efficiently utilize the potential of existing data resources by applying AL and class balancing.
The findings pave the way for better fault classification, fewer production interruptions, fewer customer claims, and higher-quality products—essential for maintaining competitiveness in the industrial fiber manufacturing industry with lower operational costs for data labeling and training classification models.

Looking ahead, it is clear that our results set the stage for a more thoughtful and adaptive approach to multiclass classification, both in the context of fiber manufacture and in more general industrial settings.
Further investigation and adaptation of the active learning and class balancing concepts to other production contexts promise increased productivity, lower costs, and better product quality.
\chapter{Conclusion and Future Works}

First and foremost, the goal of our research was to lower the quantity of labeled data needed for time series classification and the associated expenses and efforts. 
Our proposed AL framework, which is discussed in depth in Chapters 3 and 4, successfully addresses this problem. 
We observe a notable reduction in the quantity of labeled data needed to outperform the classification benchmark in the context of both tactile robotics and real-world industrial manufacturing.
Chapter 3's discussion of temporal feature generation for tactile robotics makes clear the sliding window method and its outcomes regarding multiple window times and overlap percentages.

Furthermore, we have achieved promising results in addressing the problem of class imbalance in time series datasets. 
Significant gains were achieved by adding a class balancing instance selection algorithm to the AL strategies, as is described in detail in Chapters 3 and 4 for tactile robotics and fiber manufacturing, respectively. 
This integration led us to conclude that both the AL strategies UNC and QBC can be robust in efficiently querying instances for labeling, especially in cases of highly imbalanced distributions.

Moreover, our existing results serve as a stepping stone for further exploration and enhancement of our AL framework.
A crucial future direction is to extend our work into additional real-life domains, especially those where dynamic environments pose challenges for traditional classification techniques. 
Expanding beyond the UNC, QBC, and EMC strategies, we aim to integrate and evaluate additional active learning methodologies to evaluate their performance in the context of time series classification.

\phantomsection

\bibliography{papers} 

\end{document}